%% file: uet_thesis.tex
\documentclass[a4paper, 12pt, oneside]{uet_thesis}  % Use the "Thesis" style, based on the ECS Thesis style by Steve Gunn
\graphicspath{{Figures/}}  % Location of the graphics files (set up for graphics to be in PDF format)

\usepackage[square, numbers, comma, sort&compress]{natbib}  % Use the "Natbib" style for the references in the Bibliography
\usepackage{verbatim}  % Needed for the "comment" environment to make LaTeX comments
\usepackage{makecell}
\usepackage{longtable}
\usepackage{xltabular}
\usepackage{placeins}
\usepackage{subfigure}
\usepackage{amsmath}

\usepackage{url}
\usepackage{natbib}

\hypersetup{urlcolor=blue, colorlinks=true}  % Colours hyperlinks in blue, but this can be distracting if there are many links.

\usepackage[compact]{titlesec}
\titlespacing{\section}{0pt}{*0}{*0}
\titlespacing{\subsection}{0pt}{*0}{*0}
\titlespacing{\subsubsection}{0pt}{*0}{*0}

\begin{document}
\frontmatter	  % Begin Roman style (i, ii, iii, iv...) page numbering

% Set up the Title Page
\title  {Clinical Decision Support System for Unani Medicine Practitioners}
\session {2019 -- 2023}
\advisor {Dr. Talha Waheed Sulemani}
\authors {  % please enter the students names and registration numbers
\!\!Marriyam Nadeem \quad2019-CS-5 \\
Farwa Mahmood \quad \;\;\;2019-CS-14 \\
Noor Fatima \quad \quad \quad \quad \!\!2019-CS-18 \\
\;Haider Sultan \;\;\;\;\;\;\;\;\;\;\;\;2019-CS-36 
}

\addresses  {\deptname \\ \univname}  % Do not change this here, instead these must be set in the "Thesis.cls" file, please look through it instead
\date       {\today}
\subject    {}
\keywords   {}

\maketitle
%% ----------------------------------------------------------------

\setstretch{1.3}  % It is better to have smaller font and larger line spacing than the other way round

% Define the page headers using the FancyHdr package and set up for one-sided printing
\fancyhead{}  % Clears all page headers and footers
\rhead{\thepage}  % Sets the right side header to show the page number
\lhead{}  % Clears the left side page header

\pagestyle{fancy}  % Finally, use the "fancy" page style to implement the FancyHdr headers

%% Select only one of the certification pages  
%\CertificationMSc{}
% \CertificationBSc{}
\clearpage  % Certification ended, now start a new page

%% ----------------------------------------------------------------
% Declaration Page required for the Thesis, your institution may give you a different text to place here
\Declaration{
%\addtocontents{toc}{\vspace{1em}}  % Add a gap in the Contents, for aesthetics

We declare that the work contained in this thesis is our own, except where explicitly stated otherwise. In addition, this work has not been submitted to obtain another degree or professional qualification.

\bigskip

Signed:~~ \rule[0em]{10em}{1.0pt} \\ % This prints a line for the signature 
Date:~~~~ \rule[0em]{10em}{1.0pt}  % This prints a line to write the date
}
\clearpage     % Declaration ended, now start a new page

%% ----------------------------------------------------------------

\setstretch{1.3}  % Reset the line-spacing to 1.3 for body text (if it has changed)

% The Acknowledgements page, for thanking everyone
\acknowledgements{
%\addtocontents{toc}{\vspace{1em}}  % Add a gap in the Contents, for aesthetics

First and foremost, We would like to praise Allah the Almighty, the Most Gracious, and the Most Merciful for His blessing given to us during this project and in completing this thesis. \\
\\ We would like to express sincere gratitude to our project advisor, Dr. Talha Waheed Sulemani, for their invaluable guidance and support throughout this project. Their expertise in both of the disciplines Computer Science and Unani Medicines and his feedback were instrumental in shaping the direction and scope of this work. This project would not have been possible without his expertise and dedication.\\
\\ We are also deeply grateful to our family and friends for their encouragement and constant support. Their patience and understanding during the many long hours We spent on this project were invaluable. Finally, we would like to express our appreciation to all those who have supported us in countless ways throughout this project. Your encouragement, motivation, and assistance have been vital to our success, and we could not have accomplished this without you. Thank you all for being a part of this journey.}
\clearpage  % End of the Acknowledgements

%% ----------------------------------------------------------------
% End of the pre-able, contents and lists of things
% Begin the Dedication page
\setstretch{1.3}  % Return the line spacing back to 1.3
\pagestyle{empty}  % Page style needs to be empty for this page
\dedicatory{

To Our Beloved Parents,\\
We dedicate this project to you, with heartfelt appreciation for your endless love, encouragement, and unwavering support. Your constant guidance and motivation have been the driving force behind our success, and we are forever grateful for all that you have done. This project is a testament to the values and principles that you have instilled in us, and we hope to make you proud with all that we achieve in the future. \\Thank you for everything.\\
}

%% ----------------------------------------------------------------
\pagestyle{fancy}  %The page style headers have been "empty" all this time, now use the "fancy" headers as defined before to bring them back

%% ----------------------------------------------------------------
\lhead{\emph{Contents}}  % Set the left side page header to "Contents"
\tableofcontents  % Write out the Table of Contents

%% ----------------------------------------------------------------
\lhead{\emph{List of Fig.s}}  % Set the left side page header to "List if Figures"
\listoffigures  % Write out the List of Figures

%% ----------------------------------------------------------------
\lhead{\emph{List of Tables}}  % Set the left side page header to "List of Tables"
\listoftables  % Write out the List of Tables

%% ----------------------------------------------------------------
\setstretch{1.5}  % Set the line spacing to 1.5, this makes the following tables easier to read
\clearpage  % Start a new page
\lhead{\emph{Abbreviations}}  % Set the left side page header to "Abbreviations"
\listofsymbols{ll}  % Include a list of Abbreviations (a table of two columns)
{
% \textbf{Acronym} & \textbf{W}hat (it) \textbf{S}tands \textbf{F}or \\
\textbf{AIML} & \textbf{A}rtifical \textbf{I}ntelligence \& \textbf{M}achine \textbf{L}earning\\
\textbf{API} & \textbf{A}pplication \textbf{P}rogramming \textbf{I}nterface\\
\textbf{BERT} & \textbf{B}idirectional \textbf{E}ncoder \textbf{R}epresentations \\ from \textbf{T}ransformers \\
\textbf{CDSS} & \textbf{C}linical \textbf{D}ecision \textbf{S}upport \textbf{S}ystem \\
\textbf{EHR} & \textbf{E}lectronic  \textbf{H}ealthcare \textbf{R}ecords \\
\textbf{GPT} & \textbf{G}enerative \textbf{P}retrained \textbf{T}ransformers \\
\textbf{ICL} & \textbf{I}n-\textbf{C}ontext \textbf{L}earning \\
\textbf{LLMs} &\textbf{L}arge \textbf{L}anguage \textbf{M}odels \\
\textbf{LMs} & \textbf{L}anaguage \textbf{M}odels \\
\textbf{NLP}&\textbf{N}atural \textbf{L}anguage \textbf{P}rocessing \\
\textbf{RDBMS} & \textbf{R}elational \textbf{D}atabase \textbf{M}anagement \textbf{S}ystem \\
\textbf{TCM}& \textbf{T}raditional \textbf{C}hinese \textbf{M}edicines\\
\textbf{UM}& \textbf{U}nani \textbf{M}edicines

}

%% ----------------------------------------------------------------
% The Abstract Page
\addtotoc{Abstract}  % Add the "Abstract" page entry to the Contents
\abstract{
%\addtocontents{toc}{\vspace{1em}}  % Add a gap in the Contents, for aesthetics
Like other fields of Traditional Medicines, Unani Medicines have been found as an effective medical practice for ages. It is still widely used in the subcontinent, particularly in Pakistan and India. However, Unani Medicines Practitioners are lacking modern IT applications in their everyday clinical practices. The process of diagnosing diseases can be difficult, time-consuming and prone to error. An Online Clinical Decision Support System may address this challenge to assist apprentice Unani Medicines practitioners in their diagnostic processes.\\
\\ The proposed system provides a web-based interface to enter the patient’s symptoms, which are then automatically analyzed by our system to generate a list of probable diseases. The system allows practitioners to choose the most likely disease and inform patients about the associated treatment options remotely. \\
\\ The system consists of three modules: an Online Clinical Decision Support System, an Artificial Intelligence Inference Engine, and a comprehensive Unani Medicines Database. The system employs advanced AI  techniques such as Decision Trees, Deep Learning, and Natural Language Processing. For system development, the project team used a technology stack that includes React, FastAPI, and MySQL. Data and functionality of the application is exposed using APIs for integration and extension with similar domain applications. \\
\\The novelty of the project is that it addresses the challenge of diagnosing diseases accurately and efficiently in the context of Unani Medicines principles. By leveraging the power of technology, the proposed Clinical Decision Support System has the potential to ease access to healthcare services and information, reduce cost, boost practitioner and patient satisfaction, improve the speed and accuracy of the diagnostic process, and provide effective treatments remotely. The application will be useful for Unani Medicines Practitioners, Patients, Government Drug Regulators, Software Developers, and Medical Researchers.
 
}
\clearpage  % Abstract ended, start a new page

%% ----------------------------------------------------------------
\mainmatter	  % Begin normal, numeric (1,2,3...) page numbering
\pagestyle{fancy}  % Return the page headers back to the "fancy" style
\onehalfspacing
% Include the chapters of the thesis, as separate files
% Just uncomment the lines as you write the chapters

\input{./Chapters/Chapter1} % Introduction 

\input{./Chapters/Chapter2} % What to Write 

\input{./Chapters/Chapter3} 

\input{./Chapters/Chapter4}

\input{./Chapters/Chapter5}

\input{./Chapters/Chapter6}

%% ----------------------------------------------------------------
% Now begin the Appendices, including them as separate files

\addtocontents{toc}{\vspace{2em}} % Add a gap in the Contents, for aesthetics

\appendix % Cue to tell LaTeX that the following 'chapters' are Appendices

\addtocontents{toc}{\vspace{2em}}  % Add a gap in the Contents, for aesthetics
\backmatter

%% ----------------------------------------------------------------
\label{References}
\lhead{\emph{References}}  % Change the left side page header to "References"

\bibliographystyle{plainnat}  % Use "unsrtnat" BibTeX style for formatting the references

\bibliography{references}  % The references information are stored in the file named "references.bib"

\end{document}

%% file: Chapters/Chapter1.tex
% Chapter 1

\chapter{Introduction}
% Write in your own chapter title
\label{Chapter1}
\lhead{Chapter 1. \emph{Introduction}} % Write in your own chapter title to set the page header
This chapter introduces the project, which aims to digitize and automate the Unani Medicines diagnostic process. The chapter highlights the objectives of the project and the need for its implementation in advancing Unani Medicines (UM). The scope and limitations, the deliverables of the project.

\section{A Brief History of Unani Medicines }
Unani Medicines is a traditional system of healing that has been practiced mostly in Asia for centuries. Its origins can be traced back to ancient Greece, where the Greek physician Hippocrates and his followers laid the foundation for the system. Over time, Unani Medicines incorporated elements from various cultures, including Persia, India, and the Arab world, evolving into a comprehensive medical system. UM emphasizes the balance (known as homeostasis) between the body's four elements (earth, air, water, fire) and focuses on promoting holistic well-being. It utilizes natural remedies, such as herbs, minerals, and animal products, as well as dietary and lifestyle modifications, to prevent and treat diseases \cite{parveen2022traditional}.
\\ Unani Medicines's importance lies in its holistic approach towards body and soul, natural ingredients, personalized medicines according to unique temperament of each patient, and emphasis on maintaining health rather than treating disease. All of the above mentioned features of UM make it relevant and beneficial in today's healthcare landscape. UM strives to discover the most effective ways that a person can live a long and healthy life with no or minimal disease. The experts of this knowledge believe that illnesses are prevented through the use of pure and clean water, breathing fresh air, and eating healthy food. Also, a healthy balance must be maintained between mind and body in order that the metabolic processes be performed with ease and the body is able to eliminate waste. Unani Medicines offers a belief system to promote wellness, prevent diseases, and treatments through natural drugs, diet, surgery and lifestyle changes.
It continues to be practiced and valued in many Asian countries, contributing to their healthcare system. Its knowledge and practices need to be preserved for cultural heritage and for the promotion of alternative medical options, as advised by the World Health Organization \cite{who2013traditional}.

\section{Clinical Decision Support System}
A Clinical Decision Support System (CDSS) is an IT-based solution designed to assist healthcare practitioners in making informed clinical decisions. It integrates patient-specific information with medical knowledge and provides tailored recommendations, alerts, and suggestions at the point of healthcare. CDSS has been built using rule-based, machine learning, and/or AI-based approaches, to analyze patients’ data, interpret clinical guidelines, and generate treatment recommendations. By leveraging CDSS, healthcare practitioners can ease access, reduce cost, improve the accuracy and efficiency of diagnosis, enhance treatment planning, and reduce medical errors. CDSS can support healthcare practitioners by providing access to relevant medical literature, patient histories and medical records, drug databases, and decision algorithms. They empower them with valuable insights and improve patient outcomes and satisfaction. This powerful tool of CDSS has the potential to revolutionize healthcare delivery by augmenting clinical expertise and promoting evidence-based practice \cite{sharma2021review}.

\section{Objectives of the Project}
UM has been subject to skepticism regarding its scientific validity despite having several published scientific papers on the subject \cite{itrat2020methods}. However, the incorporation of Information technology, AI, and the use of evidence-based methods can strengthen the scientific base of UM.
\\The project serves as a centralized platform for Unani Medicines practitioners, patients, researchers, herbal manufacturers, government regulatory bodies, and students, providing them easy access to various digital resources of UM in the future. By utilizing our digital resources of UM database, Inference Engine, and CDSS, software developers along with domain experts of UM can develop further knowledge-based applications for this field; e.g Electronic Health Records (EHR), knowledge portals and UM databases of herbs, drugs formulations, to name a few\cite{khan2022role}.

\section{Research Gap}
The lack of digitization of Unani Medicines practices in Pakistan has limited the ability to advance the field technologically and scientifically. Without a comprehensive AI-based IT solution specifically tailored to Unani Medicines' specific requirements \cite{waheed2020collaborative}, research and development have been hindered. While some related research has been conducted \cite{teufel2021clinical}, there is currently no UM-specific CDSS available to provide such a platform for the diagnosis and treatment of diseases. Currently UM practitioners are using manual diagnosis and treatment which can be time-consuming and error-prone.

\section{Problem Statement}
To develop an Online Clinical Decision Support System (CDSS) for Unani Medicines Practitioners that will assist them in their diagnostic and treatment process.

\section{Social and Technical Contribution of the Project}
Unani Medicines are an integral part of Pakistan’s healthcare infrastructure 
\cite{kurji2016analysis}, although it is not widely incorporated into the national healthcare system. However, it still plays an essential role in healthcare, due to its natural ingredients, low cost, easy access and seemingly low side effect medicines \cite{xu2020developing}. Therefore, there is a pressing need to build a centralized IT-based platform to support different stakeholders like Unani Medicines practitioners, patients, students, and government regulators. This platform will facilitate computational and clinical research in the field using cutting-edge Information technology tools and AI \& Machine Learning (AIML) approaches. The availability of unified UM databases and open-source AIML-based models will enable software developers to develop IT applications for Unani Medicines practitioners that will aid in diagnosis and also enable government health regulators to monitor the practices. Additionally, it will increase awareness about the historical and intellectual aspects of ancient Unani Medicines practices.

\section{Scope and Limitations of the Project}
The system will be a prototype application with limited data covering 70 diseases and their treatment options only, in English or Roman Urdu \cite{ccrum2014standard}. By any means, it is not an exhaustive effort to include each and every disease, symptom, and treatment method related to Unani Medicines. The target audience of the prototype application is Pakistanis. \\
The primary stakeholders of this project are software developers, knowledge engineers, patients, and Unani Medicines practitioners. They require easy access to all the necessary UM knowledge and resources in digital form to advance the field technologically and through cutting-edge AIML models and algorithms. Other stakeholders of the system include students, General Medical Practitioners, Herbal Drug Manufacturers, and government healthcare regulators. All of these advancements will benefit different UM stakeholders of UM and contribute to preserving the knowledge and advancing the field of Unani Medicines.

\section{The Modular Design of the Project }
The project is conceived as three interrelated modules (a UM Database, an AI Inference Engine, and a CDSS) each serving a different purpose (see Fig, \ref{fig:h_arc}, for more details about the architecture and design of each module see sections \ref{sec:arc}). That aims at digitizing and automating the whole UM diagnostic process.

    \subsection{Unani Medicines Database} 
   The first module focuses on the digitization of the UM diagnostic process that is currently confined to books and oral practices. This encompasses the development of a comprehensive graph database that stores UM data about principles, diseases, symptoms, and treatments. This will be a crucial step towards preserving and protecting the valuable but diminishing knowledge of Unani Medicines. This database will be accessible through an open-source API. Through this API, software developers and knowledge engineers can contribute to the development of this database, ensuring continuous evolution of its knowledge and improvement in functionality.
   \subsection{AI Inference Engine}
   The second module involves the implementation of an Artificial Intelligence (AI) Inference Engine. This Inference Engine offers various machine learning models that software developers and knowledge engineers can utilize and contribute to research and development for Unani Medicines digitization. Furthermore, this module enables software developers to use the functionality of AI Inference Engine through an API to build their own knowledge-based application development.
   \subsection{Online CDSS for UM Practitioners}
    The third module entails the development of an online CDSS for UM practitioners (a.k.a Hakeem) and it will demonstrate the potential applications of the UM database and AI Inference Engine modules for UM practitioners. This system utilizes both modules to assist UM practitioners in the process of diagnosing diseases based on patient symptoms. And practitioners can further recommend treatments to patients remotely. Initially, this module will be a prototype application and serves as a proof of concept for the benefits and helpfulness of automating Unani Medicines processes, showcasing how AIML-based solutions can improve patient care and outcomes in this field. Clinical decision support systems (CDSSs) are emerging solutions aimed at effectively managing and delivering extensive healthcare data to healthcare providers, with the goal of enhancing diagnosis and treatment processes. While there are promising examples of CDSS implementations in various domains, the comprehensive evidence supporting their overall benefits is still limited \cite{teufel2021clinical}.

\section{Organization of Thesis}
The organization of this thesis is as follows:\\
Chapter \ref{Chapter2}, delves into the literature review, examining the efforts to build CDSS in other domains of Traditional Medicines. Chapter \ref{Chapter 3} presents the development requirements,  methodology and environment  employed in developing this prototype application. Chapter \ref{Chapter 4} proposed a solution, consisting of the UM Database, AI Inference Engine and the CDSS modules. The system integrates modern information technology with UM for efficient diagnostic process management and personalized treatment recommendations. In Chapter \ref{Chapter 5}, the focus is on the CDSS system design, exploring its components, modules, and technologies. Chapter \ref{Chapter 6} addresses the challenge of insufficient data for the CDSS and explores various AIML techniques to enhance data availability and improve system efficiency and  concludes the study, highlighting potential areas for future research.

%% file: Chapters/Chapter2.tex
% Chapter 2

\chapter{Literature Review} % Write in your own chapter title
\label{Chapter2}
\lhead{Chapter 2. \emph{Literature Review}}
Clinical decision support systems (CDSS) play a crucial role to assist medical practitioners in making informed clinical decisions. These systems utilize patient-specific information from an Electronic Health Record (EHR) to develop algorithms that provide diagnostic assessments and treatment recommendations. In this chapter, we review a few existing efforts to develop CDSS in different disciplines of Traditional Medicines, by any means this is not an exhaustive literature review.

\section{Related Work}
\subsection{CDSS for Indian Ayurvedic Medicine}
One potential approach to personalizing the CDSS is by using Prakriti as a starting point. Prakriti and Doshas, which are humors, are closely interconnected, and the latter determines the former for each individual. In Ayurveda, it is believed that a balanced Doshas leads to good health and wellbeing, while an imbalance can lead to illness. Therefore, identifying and analyzing a person’s Prakriti using CDSS can undoubtedly aid in making informed medical decisions and help develop a specific therapeutic plan.\cite{samal2013developing}.\\
Clinical decision support systems (CDSS) are becoming more popular in traditional medicine, and more practitioners are looking to incorporate them into their practices. AyuSoft, is an online clinical decision support tool for ayurveda. The technology was created to give Ayurvedic practitioners access to precise and pertinent patient data so they could make well informed decisions. The CDSS was created with the help of professional practitioners and an open source platform with a rule based expert system. The system encompasses a wide range of Ayurveda concepts and offers direction for disease management, diagnosis, and therapy. The study emphasizes the advantages of incorporating CDSS into conventional medicine and the possibility of enhancing patient outcomes. It also highlights the difficulties involved in creating such systems and the requirement for ongoing research and development to increase their efficiency.\cite{acharya2019online}.\\
\subsection{CDSS for Homeopathy}
With the popularity of homeopathy increasing in recent years, there is an increased demand for more sophisticated instruments to support clinical judgment. Researchers have created an online clinical decision support system for homeopathy in answer to this demand. The system leverages a rule-based reasoning engine to suggest remedies based on symptoms and other patient information and uses an ontology-based method to organize knowledge on homeopathic remedies and diseases. The system is made to be simple to use and offers clinicians access to a large database of treatments as well as the option to enter patient data for analysis \cite{raghavendra2020design}. \\
\subsection{CDSS for Japanese Kampo Medicines}
Katayama, et. al. \cite{katayama2014analysis} have worked on a decision support system for Japanese medicine. The paper focuses on the "medical interview" in Japanese traditional medicine, which incorporates a questionnaire about the patient's lifestyle and subjective symptoms. They conducted an analysis of the "medical interview" data to establish an indicator for non-Kampo specialists without technical knowledge to perform suitable traditional medicine. They employed the random forests algorithm, a powerful classification algorithm, to predict the Sho.Their strengths are Comprehensive analysis as the paper thoroughly examines the "medical interview" in Kampo medicine, providing insights into its importance in Japanese traditional medicine, Practical application as the study develops an indicator for non-Kampo specialists to utilize traditional medicine effectively using data analysis and classification techniques, and The inclusion of BMI data in the "medical interview" enhances the accuracy of predicting "Sho," showcasing the researchers' commitment to improving the decision support system. Their limitations are limited discriminant ratio on test data, Reliance on a single algorithm as it uses the random forests algorithm and lack of discussion on confounding factors that may affect prediction accuracy, compromising the validity and reliability of the findings.\\
\subsection{CDSS for Korean Oriental Medicines}
Sung HK et. al. \cite{sung2019trends} have worked on a decision support system for korean medicine. Their focus is on establishing a Clinical Decision Support System (CDSS) in Korean medical services with a specific emphasis on herbal medicine prescription. They used analysis techniques of the current prescription practices of Traditional Korean Medicine doctors and investigated the implementation of CDSS through a questionnaire survey. Their strengths are that they implement a CDSS in Traditional Korean Medicine that enhances standardization, It analyzes current prescription practices and gathers insights from doctors through a survey and It also examines CDSS implementation among software developers, providing practical perspectives. Their limitations are standardizing clinical information and prescription practices pose challenges for implementing a CDSS in Traditional Korean Medicine and Variations in prescription support functions among clinics and hospitals may hinder uniformity. \\
\subsection{CDSS for Traditional Thai Medicines}
Jumphaew, et. al. \cite{jumphaew2015development} have worked on decision support systems for Traditional Thai Medicine. Their focus is on creating a prototype model for a decision-making support tool in Thai traditional medicine. They utilize data mining techniques, specifically classification algorithms such as J48, RandomTree, and HoeffdingTree. Additionally, the researchers used the Weka program (version 3.9.5) for developing the classifier model. The strengths include the high accuracy achieved by the classifier model developed from the RandomTree algorithm, with an average accuracy of 99.71\%. The precision, recall, and overall F-measure efficiency of the model were also high, indicating reliable predictions. The use of data mining techniques and the application of a forecasting model contribute to the effectiveness of the clinical decision-making support tool in Thai traditional medicine.Their limitations include the limited availability of comprehensive and up-to-date data on Thai traditional medicine. This could pose challenges in accurately training and validating the prediction model, as well as in ensuring the tool's effectiveness across different patient populations and evolving healthcare practices. Additionally, the reliance on data mining techniques may introduce potential biases or limitations in the representation and interpretation of medical information. Further research and refinement of the tool's algorithms and data sources would be necessary to address these limitations and enhance its overall performance.
\subsection{CDSS for Traditional Chinese Medicines}
Xuezhong Zhou, et. al. \cite{xuezhong2010development} have worked on a decision support system for chinese medicine. Their focus is on developing a clinical data warehouse (CDW) system for TCM to enable medical knowledge discovery and TCM clinical decision support (CDS) . They used a clinical reference information model (RIM) and physical data model for managing TCM clinical data. They have implemented an extraction-transformation-loading (ETL) tool for data integration and normalization. The CDW incorporates online analytical processing (OLAP) and complex network analysis (CNA) components. Data mining methods, including support vector machine, decision tree, Bayesian network, association rule, and CNA, are utilized for knowledge discovery. Their strengths are  that  they integrate a large volume of TCM inpatient and outpatient data, enabling multidimensional analysis and case-based decision support and successfully conducted various data mining applications. Their limitations are in terms of data quality, generalizability to larger populations, integration challenges with diverse data sources, interpretation without clinical context, and scalability to handle large data volumes and future growth.
\\

Zhang, W., \& Chen, S. \cite{zhang2009equilibrium} have worked on decision support system for Chinese medicine. Their focus is on developing a computational model of YinYang WuXing (YYWX) based on bipolar set theory. They used YinYang bipolar linear algebra and formulated dynamic equations for YYWX. The research established and proved global and local equilibrium and nonequilibrium conditions. Computer simulations demonstrated the scientific basis of the approach for herbal medicine, Qi, JingLuo, and acupuncture. Their strengths are that it introduces YinYang bipolar linear algebra and computational model of YinYang WuXing (YYWX) in TCM, Defines nourishing and regulating relations, formulates dynamic equations, and establishes equilibrium conditions, Provides a scientific basis for research in herbal medicine, Qi, QiGong, JingLuo, and acupuncture, Prototypes a decision support system for diagnostic decision analysis in TCM. Their limitations are lack of formal mathematical or scientific foundation in YYWX theory, the effectiveness and accuracy of the proposed model and decision support system need further validation, applicability of YinYang cellular network architecture to other fields requires verification.\\
 Li et. al. \cite{li2014clinical} have worked on a decision support system for Chinese medicine. Their focus is on assisting inexperienced clinicians in making reliable clinical decisions and improving the curative effectiveness in TCM. The system use integration of TCM clinical cases data from a TCM clinical data warehouse and employs CBR for retrieving similar cases. It also includes a flexible diagnosis and treatment modification mechanism based on correlation analysis. The strengths of their approach lie in utilizing TCM clinical cases and providing personalized diagnosis and treatment. Their limitation is the reliance on TCM electronic medical record data. The system's effectiveness heavily relies on the accuracy and comprehensiveness of the data within the TCM clinical data warehouse. If the data is incomplete, inconsistent, or of low quality, it may lead to inaccurate recommendations and unreliable clinical decisions. Additionally, the system's reliance on case-based reasoning (CBR) algorithm may limit its ability to handle complex and rare cases that deviate from the existing clinical cases in the dataset. This could potentially result in inadequate support for clinicians facing unique or challenging situations.\\

  Hai Long, et. al. \cite{long2020ontology} have worked on a decision support system for Chinese medicine. Their focus is on developing an ontology-based decision support system for the syndrome-differentiation and treatment of Psoriasis Vulgaris in Traditional Chinese Medicine (TCM). They used  a domain ontology for syndrome differentiation of psoriasis vulgaris using the ontology editor Protégé. They employed a top-down approach based on the framework of General Formal Ontology (GFO) and its middle-level core ontology GFOTCM. Additionally, they implemented a prototype system named ONTOPV-system, which incorporated fuzzy logic reasoning, case retrieval using Case Based Reasoning (CBR), and a fuzzy pattern recognition approach.\\Their  strengths are  Integration of Syndrome Differentiation as the system incorporates syndrome differentiation, a key aspect of TCM diagnosis, enhancing diagnostic accuracy for psoriasis vulgaris, Domain Ontology Development,  Expert-Assisted Decision Support as the system combines TCM practitioners' expertise with its functionalities, improving clinical diagnostic decisions for psoriasis vulgaris, utilizing Fuzzy Logic Reasoning enables flexible and nuanced decision-making, considering uncertainty and imprecise information,  by utilizing Case-Based Reasoning, the system learns from previous cases, providing valuable insights for diagnosis and treatment decisions and Extensible Knowledge Base. The limitations of the developed TCM decision support system for psoriasis vulgaris include reliance on accurate data, dependence on the quality of the underlying ontology, challenges in accommodating individual variations, uncertainty introduced by fuzzy logic reasoning, and the need for further validation and evaluation.\\
 
 Rakhma Oktavina, Retno Maharesi \cite{oktavina2013decision} have worked on a decision support system for traditional chinese medicines production. Their focus is  to develop a decision support system in herbs production scheduling appropriate for Good Traditional Medicine Manufacturing Practices (GTMMP). They used an algorithm for the scheduling decision support system, complying with GTMMP standards. Their system was designed using a network analysis technique combining the Evaluation and Review Technique Program (PERT) and Critical Path Method (CPM).  Their strengths are  Tailored decision support systems for GTMMP requirements, enhancing production scheduling in TCM manufacturing. Comprehensive network analysis technique (PERT and CPM) used to design the scheduling algorithm and inclusion of efficient database and model management systems for data and model organization. Their major issue is not supporting TCM practitioners in the Diagnostic process. Other limitations are Limited generalizability, Lack of validation and Scalability challenges.\\

\begin{table}[h!]
\centering
\resizebox{\textwidth}{!}{%
\begin{tabular}{|p{1.8cm}|c|p{2.5cm}|p{3cm}|p{4cm}|p{4cm}|p{4cm}|p{4cm}|}
\hline
\textbf{Reference} & \textbf{Year} & \textbf{TM domain} & \textbf{Focus} & \textbf{Technologies/Tools} & \textbf{Results} & \textbf{Strengths} & \textbf{Limitation} \\
\hline
\cite{katayama2014analysis}& 
2014 & 
Japanese Medicine & 
"Medical interview" which incorporates a questionnaire about the patient's lifestyle and subjective symptoms. & 
The authors conducted an analysis of the "medical interview" data to establish an indicator for non-Kampo specialists without technical knowledge to perform suitable traditional medicine. They employed the random forests algorithm, a powerful classification algorithm, to predict the Sho.& Research confirms random forests' predictive power for Kampo medicine diagnosis using "medical interview" data. After cleaning, discriminant ratio increased from 67.0\% to 72.4\% Addition of BMI raised ratio to 91.2\%, emphasizing BMI's key role in classification. & 
Strengths include comprehensive analysis of "medical interview" in Kampo medicine, practical indicator development for non-specialists using data techniques, and enhanced prediction accuracy of "Sho" diagnosis through BMI data integration, reflecting dedication to decision support system enhancement. &
Standardizing clinical information and prescription practices pose challenges for implementing a CDSS in Traditional Korean Medicine and Variations in prescription support functions among clinics and hospitals may hinder uniformity. \\
\hline
\cite{xuezhong2010development} & 
2010 & 
Chinese Medicine & 
Developing a clinical data warehouse system to enable medical knowledge discovery and CDSS. & 
They used clinical models for TCM data, developed an ETL tool for integration and normalization, and employed diverse data mining methods for knowledge discovery. The CDW includes OLAP and CNA components. & 
CDW incorporates vast TCM clinical data, facilitating multifaceted analysis and case-driven decision support. Data mining extracts insights on syndrome differentiation, acupuncture points, and herb combinations. & 
Their strengths are  that  they integrate a large volume of TCM inpatient and outpatient data, enabling multidimensional analysis and case-based decision support and successfully conducted various data mining applications. & 
Their limitations are in terms of data quality, generalizability to larger populations, integration challenges with diverse data sources, interpretation without clinical context, and scalability to handle larger data volumes and future growth. \\
\hline
\cite{sung2019trends} & 
2019 & 
Korean Medicine & 
Establishing a CDSS with a specific emphasis on herbal medicine prescription. & 
They used analyzing techniques of the current prescription practices of Traditional Korean Medicine doctors and investigated the implementation of CDSS through a questionnaire survey.& 
In clinics, 41.2\% manipulated 1-4 herbs, while 31.2\% adjusted 4-7 herbs. In hospitals, 52.5\% adjusted 1-4 herbs, and 35.5\% adjusted 4-7 herbs. Doctors desired prescription support functions in the electronic medical record system, including medicine information, herb combinations, and search options by prescription efficacy. Implementation of these functions varied among clinics and hospitals. & 
Their strengths involve CDSS implementation in Traditional Korean Medicine for better standardization. They analyze prescriptions, gathering doctor insights via a survey. The study explores CDSS adoption by software developers, providing practical insights.& 
Their limitations involve challenges in standardizing clinical information and prescriptions for CDSS in Traditional Korean Medicine. Varied prescription support functions across clinics and hospitals may hinder uniformity.\\
\hline
\end{tabular}%
}
\end{table}
\clearpage

\begin{table}[h!]
\centering
\resizebox{\textwidth}{!}{%
\begin{tabular}{|p{1.8cm}|c|p{2.5cm}|p{3cm}|p{4cm}|p{4cm}|p{4cm}|p{4cm}|}
\hline
\textbf{Reference} & \textbf{Year} & \textbf{TM domain} & \textbf{Focus} & \textbf{Technologies/Tools} & \textbf{Results} & \textbf{Strengths} & \textbf{Limitation} \\
\hline

\cite{zhang2009equilibrium} & 
2009 & 
Chinese Medicine & 
Developing a computational model of YinYang WuXing (YYWX) based on bipolar set theory. &
They applied YinYang algebra, creating dynamic equations for YYWX. The research confirmed global and local equilibrium, using computer simulations for herbal medicine, Qi, JingLuo, and acupuncture.& 
It proposes YinYang bipolar linear algebra model for TCM, offering scientific basis, diagnostic analysis, and applications in herbal medicine, acupuncture, and biomedicine. &
Their strengths are that it introduces YinYang algebra and computational model (YYWX) in TCM, formulates dynamic equations, establishes scientific basis for herbal medicine and acupuncture research, and prototypes diagnostic decision support system.&
Their limitations are lack of formal YYWX theory foundation, need for model validation, YinYang cellular network's applicability to other fields unverified.\\
\hline
\cite{oktavina2013decision} & 
2013 & 
Herbs Traditional Medicine & 
To develop a CDSS in herbs production scheduling appropriate for Good Traditional Medicine Manufacturing Practices (GTMMP). & 
They employed a GTMMP-compliant algorithmic scheduling decision support system. Designed using a network analysis technique merging PERT and CPM. & 
The study created a GTMMP-aligned decision support system for herb production scheduling. Utilizing network analysis, it offers a database and model management. Ideal for GTMMP certification-seeking companies.&
Their strengths are Custom GTMMP-compliant decision support system, optimized production scheduling, robust network analysis (PERT and CPM), efficient database management. & 
Their limitations are  Limited generalizability, Lack of validation and Scalability challenges. \\
\hline 
\cite{jumphaew2015development}&
2015&
Thai Medicine &
Their focus is on creating a prototype model for a decision-making support tool in Thai traditional medicine. &
They utilize data mining techniques, specifically classification algorithms such as J48, RandomTree, and HoeffdingTree. Additionally, the researchers used the Weka program (version 3.9.5) for developing the classifier model. &
The research produced a robust clinical decision support tool for Thai traditional medicine with a 99.71\% average accuracy in prediction. The web application prototype can aid practitioners in informed treatment decisions.&
The strengths include the high accuracy achieved by the classifier model developed from the RandomTree algorithm, with an average accuracy of 99.71\%. The precision, recall, and overall F-measure efficiency of the model were also high, indicating reliable predictions. The use of data mining techniques and the application of a forecasting model contribute to the effectiveness of the clinical decision-making support tool in Thai traditional medicine.&
The scarce, outdated data on Thai traditional medicine hampers accurate model training and validation. This undermines cross-population efficacy and adapting to evolving healthcare trends. Data mining reliance might introduce biases and data representation constraints. Further research is required to refine algorithms and data sources for improved tool performance.\\
\hline
 \cite{li2014clinical} &
2014 & 
Chinese Medicine &
Their focus is on assisting inexperienced clinicians in making reliable clinical decisions and improving the curative effectiveness in TCM. &
The system use integration of TCM clinical cases data from a TCM clinical data warehouse and employs CBR for retrieving similar cases. It also includes a flexible diagnosis and treatment modification mechanism based on correlation analysis. &
The study proves the effectiveness of a TCM clinical decision support system. By integrating cases and using case-based reasoning, the system aids novice clinicians, enhancing treatment decisions and outcomes. It adapts to personalized treatment, boosting practicality in real clinical scenarios.&
The strengths of their approach lie in utilizing TCM clinical cases and providing personalized diagnosis and treatment.  &
Limitations include dependence on TCM electronic medical record data quality, potentially leading to inaccurate recommendations. Relying on case-based reasoning might hinder handling complex or rare cases.\\
\hline
\cite{long2020ontology} &
2020&
Chinese Medicine &
Developing an ontology-based decision support system for the syndrome-differentiation and treatment of Psoriasis Vulgaris in Traditional Chinese Medicine (TCM). &
They utilized domain ontology for psoriasis vulgaris differentiation via Protégé editor. A top-down approach using General Formal Ontology (GFO) framework and GFOTCM core ontology was employed. An ONTOPV-system prototype was developed, integrating fuzzy logic reasoning, Case Based Reasoning (CBR), and fuzzy pattern recognition.& 
The study developed a prototype system for TCM syndrome differentiation in psoriasis diagnosis. It uses domain ontology, fuzzy logic, and case retrieval, offering clinical decision support for TCM practitioners and showing promising potential.&
Their strengths include Syndrome Differentiation integration for better psoriasis vulgaris diagnosis, Domain Ontology Development, Expert-Assisted Decision Support, Fuzzy Logic Reasoning for adaptable decisions, Case-Based Reasoning for insights from prior cases, and an Extensible Knowledge Base.&
The limitations of the developed TCM decision support system for psoriasis vulgaris include reliance on accurate data, dependence on the quality of the underlying ontology, challenges in accommodating individual variations, the uncertainty introduced by fuzzy logic reasoning, and the need for further validation and evaluation.\\
\hline
\end{tabular}%
}

\caption{CDSS Literature Review Matrix}
\label{tab:lit_review_matrix}
\end{table}

\section{Relevant Applications}
\subsection{Isabel} Isabel is a diagnostic decision support system used by healthcare professionals worldwide. It helps doctors in the diagnosis of complex medical conditions by analyzing patient symptoms and medical history. The system generates a list of potential diagnoses and provides relevant clinical information to aid in the decision-making process.
\subsection{VisualDx} VisualDx is a CDSS that assists healthcare providers in diagnosing visually identifiable diseases. It uses a vast database of medical images and associated clinical information to support differential diagnoses. The system helps physicians compare patient symptoms with visual representations to arrive at accurate diagnoses.
\subsection{Ada Health}Ada Health is an AI-powered symptom assessment tool that allows users to input their symptoms and receive a personalized analysis of potential conditions. The app uses machine learning algorithms to provide insights into possible diagnoses and offers recommendations for further medical evaluation.
\subsection{Buoy Health}Buoy Health is an AI driven chatbot that helps users assess their symptoms and provides recommendations for appropriate healthcare actions. It uses natural language processing and machine learning techniques to simulate a conversation with users, guiding them toward potential diagnoses and suggesting next steps for care.
\subsection{UpToDate} UpToDate is an evidence based clinical decision support resource widely used by healthcare professionals. It provides access to a comprehensive collection of medical information, including disease summaries, treatment guidelines, and expert recommendations. UpToDate assists clinicians in staying updated with the latest medical knowledge and aids in making informed decisions.
\subsection{Infermedica}Infermedica offers an AI driven symptom checker and triage platform that assists both patients and healthcare providers in the diagnostic process. The system collects information about symptoms and medical history, analyzes the data, and generates potential diagnoses along with recommendations for further evaluation or treatment.
\subsection{Medscape}Medscape is a popular medical information platform that provides access to a wide range of clinical resources, including medical news, drug information, and professional education materials. It serves as a valuable reference tool for healthcare professionals in various specialties, supporting them in decision-making and staying informed about medical advancements.
\subsection{Epocrates} Epocrates is a mobile app designed for healthcare professionals, offering clinical decision support tools and drug information resources. It provides quick access to medical reference materials, drug interactions, dosing guidelines, and clinical calculators, assisting healthcare providers in making accurate and efficient decisions.
\subsection{Mayo Clinic Symptom Checker} The Mayo Clinic Symptom Checker is an online tool that enables users to enter their symptoms and receive a list of potential conditions to consider. It offers reliable information based on Mayo Clinic's medical expertise and helps users understand the possible causes of their symptoms.
\subsection{Babylon Health} Babylon Health is a digital healthcare platform that includes a symptom checker and virtual consultations with healthcare professionals. The platform uses AI algorithms to analyze symptoms and provide potential diagnoses. It also offers video consultations for patients to interact with doctors remotely.

\begin{table}[h]
\centering
\begin{tabularx}{\textwidth}{|X|X|X|}
\hline
 \textbf{Name} & \textbf{Purpose} & \textbf{Limitation} \\
\hline
Isabel & 
Aids medical professionals in diagnosing conditions with potential diagnoses and clinical data. & 
Depends on accurate input and user data, may not include rare or emerging diseases, not a substitute for healthcare professional expertise.\\
\hline
VisualDx & 
Helps healthcare providers visually identify diseases using a large medical image database and related clinical information. &
VisualDx accuracy depends on image database quality, may not cover all conditions, and could overlook non-visual symptoms or underlying issues. \\
\hline
Ada Health & 
Offers personalized symptom assessments and potential diagnoses to empower users in making health decisions. &
Ada Health's insights use algorithms and medical knowledge, not a substitute for medical guidance. Consult a healthcare professional for accurate diagnosis and treatment.\\
\hline
Buoy Health & 
Provides an AI chatbot for symptom assessment and offers guidance on suitable healthcare actions. &
Chatbot provides algorithm-based responses, not a substitute for thorough medical exams. Use for initial guidance, consult a professional for comprehensive advice. \\
\hline
UpToDate & 
Offers healthcare professionals current medical knowledge, guidelines, and expert insights for clinical decision-making. &
UpToDate provides valuable information but should complement clinical expertise, as recommendations may differ based on patient characteristics or local guidelines.\\
\hline
Infermedica & 
The symptom checker helps users self-assess symptoms and suggests potential diagnoses for further evaluation. &
The symptom checker helps users self-assess symptoms and suggests potential diagnoses for further evaluation.\\
\hline
\end{tabularx}
\end{table}
\clearpage

\begin{table}
\centering
\begin{tabularx}{\textwidth}{|X|X|X|}
\hline
 \textbf{Name} & \textbf{Purpose} & \textbf{Limitation} \\
\hline
Medscape & 
The symptom checker helps users self-assess symptoms and suggests potential diagnoses for further evaluation. &
Medscape offers reference and education, not a substitute for personalized clinical judgment or professional consultation.\\
\hline

Epocrates & 
Provides clinical decision support tools, drug info, and medical references for healthcare professionals.&
Epocrates are helpful but combine with clinical judgment and current guidelines.\\
\hline
Mayo Clinic Symptom Checker & 
Assists in identifying symptom causes and offers general information. &
The symptom checker provides general guidance, not a definite diagnosis, and should not replace medical consultation. \\
\hline
Babylon Health & 
Offers digital platform with symptom checker and virtual consultations for convenient healthcare access. &
Virtual consultations have limitations for physical examination and may not be suitable for all conditions. Consider patient needs and seek in-person care when necessary.\\
\hline
\end{tabularx}
\caption{CDSS Applications Comparison Matrix}
\label{tab:acm}
\end{table}
We can see from the above CDSS apps, there exists no single app that has been developed according to Unani Medicines principles and offers UM drugs and treatment options. So there is a need to develop such an app that supports UM practitioners in their everyday diagnostic process.
\section{Research Gap}
 Based on the above literature and apps review (Table \ref{tab:lit_review_matrix}, Table \ref{tab:acm}), it is evident that there exist no CDSS specifically designed for Unani Medicines. Similar to other traditional medicines disciplines, practitioners of UM are in need of a CDSS to enhance their everyday medical practices. In our project, we aim to address this gap by developing a comprehensive CDSS tailored to the specific requirements and diagnostic approaches of Unani Medicine. This system will provide valuable support and guidance to Unani Medicine practitioners, ultimately improving the quality of patient care and treatment outcomes.

%% file: Chapters/Chapter3.tex
\chapter{Development Requirements, Methodology and Environment} % Write in your own chapter title
\label{Chapter 3}
\lhead{Chapter 3. \emph{ Development Requirements, Methodology and Environment}}
This chapter introduces the research methodology and user requirements of the project that focuses on the development of a comprehensive platform for Unani Medicines. The chapter begins with a general description of the project, highlighting the functions of each module (see fig \ref{fig:h_arc} for details )
\section{General Description}
\subsection{Functions of Module}
\begin{enumerate}
   \item The implementation of the UM Database module will enable the digitization of the Unani Medicines, which will provide easy access to this knowledge for educational and healthcare purposes. It pioneers the digitization of Unani Medicines data, a critical step towards the preservation and accessibility of invaluable knowledge in this domain. It will help preserve the knowledge of Unani Medicines and make it available to pharma and botanical researchers, UM practitioners, and students.
    \item The AI Inference Engine module will advance research and development in the field of Unani Medicines through the use of state of the art AIML algorithms. While several such models have been developed for other traditional medicines disciplines, such as BioBERT, BioGPT, and DrBERT. The prototype application will enable software engineers to develop Language Models, such as UnaniBERT and UnaniGPT, that are specific to Unani Medicines.
 \item The Online CDSS will be a valuable resource for UM practitioners and patients. The system will standardize the diagnostics process in Unani Medicines, making it easier for practitioners to communicate with each other and patients. The CDSS will also improve patient access to healthcare information and increase practitioner and patient satisfaction. CDSS is the first of its kind, tailored specifically for Unani Medicines. While various other CDSS exist, none are designed to cater to the needs of Unani Medicines practitioners in particular.
\end{enumerate}

% \subsection{Similar System Information}
% The current project serves as an innovative breakthrough in the field of Unani Medicine in Pakistan. 
% \begin{enumerate}
%     \item It pioneers the digitization of Unani Medicines data, a critical step towards the preservation and accessibility of invaluable knowledge in this domain.
 
%     \itemIt introduces an AI Inference Engine that presents state of the art AI models for research and development in Unani Medicines. While several AI models have been developed for traditional medicine, such as BioBERT, BioGPT, and DrBERT, no UnaniBERT or UnaniGPT exists yet, making this project a trailblazer in the field.

%     \item The Clinical Decision Support System (CDSS) developed in this project is the first of its kind, tailored specifically for Unani Medicines. While various other CDSS exist, none are designed to cater to the needs of Unani Medicines practitioners.
% \end{enumerate}

\subsection{User Characteristics}
\begin{enumerate}
\item The application prototype targets a diverse group of users, including Pharma and Botanical Researchers, Software Developers, UM Students, UM Practitioners, Patients, and Government Health Regulators.
\item To utilize the UM Database API, the user must possess technical expertise in Computer Science and Software Development, including knowledge of APIs, Databases, and Ontologies.
\item For the AI Inference Engine, users must have a solid foundation in Computer Science, AIML, Data Science, NLP, and APIs to effectively leverage the state of the art AIML models available in the system.
\item As for the CDSS, users must have an in-depth understanding of Unani Medicines to fully benefit from the system’s functionalities.
\end{enumerate}

\subsection{User Problem Statement}
The lack of digitization of Unani Medicines has limited the ability to advance the field technologically. Currently manual diagnostic practices in UM are time consuming and error prone. While some computational research has been started in Unani Medicines Digitization \cite{waheed2013development} \cite{waheed2020collaborative} \cite{mehmood2023modelling} \cite{naz2021cloud} \cite{amjad2014ontology}, there is currently no specific CDSS to provide a platform for the diagnosis and treatment of diseases as per UM principles.

\subsection{User Objectives}
The user aims to access and contribute to the digitization of Unani Medicines data to enable easier and more efficient access to information for various purposes. The user seeks to participate in the research and development of state of the art AIML models in the UM to improve the quality and accuracy of diagnosis and treatment. The user desires a CDSS to facilitate the digitization of diagnosis and treatment of diseases, ensuring efficient and standardized care for patients.

\subsection{General Constraints}
The user should have easy and intuitive access to the data using the API. The process to fetch the required data should be simple and straightforward, and the process to obtain permission to contribute to the data should not be overly time consuming. Users will need to verify their contributions to ensure the accuracy and reliability of the data.\\
Access to the Inference Engine will require a strong understanding of AIML models. The process to access the models for the development of applications should be streamlined to minimize any potential delays. Software developers and knowledge engineers wishing to access the AIML models and contribute to their development will be required to provide proof of their academic background and research goals.\\
The Online CDSS is a proof of concept and should be designed with simplicity and ease of use in mind. It should be straightforward for patients and practitioners to access and navigate the system to obtain the information they need.
\section{Research Methodology}
\subsection{Solving the Problem of Little Unani Medicines Data}
This section focuses on how Artificial Intelligence \& Machine Learning (AIML) can be applied in our Online CDSS. To identify the optimal approach, different AIML methodologies such as Decision Trees, Deep Learning, and Natural Language Processing (NLP) were explored. In our exploration of AIML techniques, we ventured into distinct modeling strategies. Initially, a Decision Tree Model was employed, using clinical rules and predicates from existing UM data for the CDSS. However, its rigidity and reliance on predefined predicates limited its adaptability. Shifting to a Deep Learning Model, data scarcity was addressed through augmentation, though challenges of limited source data and declining quality arose. This led to insights into advanced Natural Language Processing (NLP) techniques. In the realm of NLP (6.2.3), despite limited data, a novel approach emerged – augmenting data quality through NLP prompts generated from engineered data. This substantially boosted data volume, enabling successful training of a BERT model (specifically, fine-tuning a BioBERT model ) for accurate classification. This comprehensive exploration underscores the evolution from Decision Trees to Deep Learning and NLP, each contributing unique perspectives and insights, ultimately enhancing the development of an effective CDSS.
\subsubsection{Source, Size and Nature of Dataset} 
Our main source of UM data was a book of ‘Standard Unani Treatment Guidelines for Common Disease (Vol-1)’ \cite{ccrum2014standard}, containing information about 70 diseases, including their symptoms and causes, treatment principles, pharmacotherapy, regimental therapy, dietary recommendations and restrictions, prevention and precautions. However, the limited information in the source book presented a hurdle for machine learning algorithms, most of them require large datasets for training. To address this difficulty, we used an innovative AIML based approach, by devising models with the ability to produce efficient results by manipulating and augmenting data using only the existing information from the source book.
\subsubsection{Future Directions - UnaniBert and UnaniGPT}
Following the solution of the little UM data issue, our next objective is to automate the system and make it more user-friendly for forthcoming software developers and knowledge engineers to digitize data and implement their own NLP techniques. To facilitate them, we have identified several  models  including GPT-3, a powerful language model with remarkable text generation and comprehension abilities; CLIP, which excels in understanding images and their associated texts; T5, a versatile model for various NLP tasks; ELECTRA, known for its efficient pre-training process; DALL-E, a model that generates creative images from textual descriptions; BERT, renowned for bidirectional context understanding; XLM-R, a multilingual model; and UniLM, specialized in generating text. These models support such efforts, including the capability to directly answer questions by parsing the UM documents, thereby eliminating the need to convert them into structured and machine-readable data.
To further support this automation, we propose the development of two novel variants of the BERT and GPT models, namely, UnaniBERT and UnaniGPT. UnaniBERT will be fine-tuned on a large corpus of UM data and will primarily focus on query-based searches and classification tasks. 
\\On the other hand, UnaniGPT will be fine-tuned on the same dataset and will possess an interactive ability to generate responses, rendering it a suitable conversational AI tool for UM. It is crucial to recognize that all of these tasks necessitate large amounts of UM data and resources. In summary, our project aims to initiate the research process in this field by rendering the UM diagnostics data more accessible and user-friendly by providing APIs, with the hope of encouraging further knowledge-based research and Apps development in the future.
Some milestones we want to achieve by the end of the project:
\begin{enumerate}
    \item UnaniBERT, fine-tune BioBERT on more data than the initial testing data to create a much stronger first version.
    \item UnaniGPT, fine-tune BioGPT (pre-trained on Bio-Medical data, by Microsoft) to create a generative model for Unani Medical Data by utilizing the generative abilities of GPT (Generative Pre-trained Transformers) architecture.
    \item UnaniT5, introduces a translation model for making the Unani Medicine data multilingual. This will be accomplished by fine-tuning Flan-T5 (by Google) a fine-tuned version of T5 (by Google).
    \item Querying NLP Models for different tasks like querying documents, querying data in tables, and querying by context.
\end{enumerate}
\subsection{AI Inference Engine}
\subsubsection{Overview}\label{section:3.2.2.1}
This module encompasses the artificial intelligence (AI) operations of the project, which required a significant amount of research. Initially, our goal was to classify diseases based on their symptoms and causes, but we encountered unforeseen complications that arose from shortage of data. As a result, our first module focused on the digitization of data, making it reliant on the success of our initial efforts.
\\To address the issue of insufficient data, we explored and investigated various techniques. Our initial attempts involved utilizing a Decision Tree Model. We carefully curated rules and predicates from the available data to support a decision making system. However, this method had limitations as it was overly static and only capable of making decisions based on a predetermined set of predicates. Furthermore, it was an inadequate solution that did not fully address our research problem and lacked scalability for future research.Our next approach to resolving the issue involved exploring a Deep Learning Model. However, due to the insufficient amount of data, we needed to engineer and augment the data before it could be used for training. To achieve this, we collected data from the source book \cite{ccrum2014standard} and employed various data engineering techniques to increase the quantity of available data. Although this approach improved the quantity of data available, we were still limited by the limited amount of data in the book. Furthermore, the quality of the data began to deteriorate as we engineered more data. Nevertheless, this exploration provided us with valuable insights that opened up new avenues for solving the problem at hand.
\\Exploring data engineering techniques in the AIML solution opened up a new avenue for us to pursue, specifically in the field of Natural Language Processing (NLP). Although it may seem counter-intuitive to utilize NLP when working with a limited amount of data, we discovered a novel approach to augment our data without sacrificing its quality. We generated NLP prompts based on the previously engineered data to augment our data effectively. Using this method, we were able to increase the amount of data to a significant level, which allowed us to train a BERT model (more specifically, fine tune a BioBERT model) on the generated sentences for classification.

\subsubsection{AI Inference Engine Research}
Two types of Decision Trees were explored. The first was developed based on rules and predicates to construct an algorithm for disease classification utilizing symptoms and causes. Essentially, this structure utilized an IF-ELSE tree format, which is inherently limited and static in its decision making capacity, unable to fully encapsulate the complexity of the problem at hand. The second type, however, demonstrated greater proficiency, as it was developed using Machine Learning methods to design the decision tree, utilizing the scikit-learn library Decision Tree. This algorithm was capable of handling much more complex and dynamic classification problems, yet it still remains relatively less advanced in comparison to other Artificial Intelligence models \cite{scikit-learn}.\\
Deep Learning is a powerful method that can handle complex classification problems effectively. However, lack of data can significantly hinder its performance. To address this issue, we employed data engineering and augmentation techniques. One approach we used involved the careful development of an algorithm that removes one of the symptoms or causes to check if the resulting vector of features can still uniquely identify the target disease. While this method helped us to increase the amount of data, it is important to note that it could potentially decrease the quality of the data over time. As we resample the new sample points based only on the already existing data, which is limited and cannot fully capture real world complexity, there could be diseases outside of our dataset that can also be classified using the same feature vectors. We have no information about these hypothetical samples before we add the newly engineered sample to our dataset.
\\If one of these hypothetical samples ends up in our dataset, the quality of the data will decrease as two identical feature vectors classify two different targets. This highlights the importance of data quality and the potential consequences of using insufficient or unreliable data for training AI models.
\\The domain of Natural Language Processing (NLP) has witnessed significant advances in the recent past with the introduction of Language Models (LMs), particularly the Large Language Models (LLMs). In our research, we have explored the capabilities of LLMs, specifically their in-context learning ability, which allows for zero shot learning if provided with sufficient context. This capability was crucial in our efforts to engineer NLP prompts by leveraging the context of symptoms and causes, using the ChatGPT language model.
To curate a large amount of data, we utilized previously engineered data from the Deep Learning method. However, this method had limitations, as the degradation of data quality was a concern, and the data was insufficient for training a language model from scratch. To overcome these challenges, we used the ChatGPT language model to generate accurate prompts that were fully in-context with the Disease, Symptoms, and Causes provided. The in-context learning ability of ChatGPT ensured the production of high quality and accurate prompts, surpassing the accuracy of human doctors in diagnosing diseases, according to a recent study. With this confidence in the data generated by ChatGPT, we introduced the first version of UnaniBERT, which is a fine-tuned BioBERT variant for Unani Medicines data.
We chose to fine-tune the BioBERT model as it was pre-trained on Bio-Medical data, making it a suitable candidate for fine-tuning with Unani Medicines data.
\subsubsection{Conclusion}
In conclusion, our rigorous journey through these modeling approaches has provided us with a comprehensive perspective on the intricacies of developing a proficient CDSS. The final stride towards Natural Language Processing showcased the potential of merging advanced techniques with carefully curated data engineering. This approach not only alleviated the challenge of data scarcity but also paved the way for more accurate diagnoses and personalized recommendations within the CDSS. As we move forward, the lessons learned from these endeavors will undoubtedly shape the evolution of our project, making it better equipped to empower healthcare professionals in making informed decisions.
\section{Development Environment Setup}
To set up the environment for our project, the following steps were undertaken:
\subsection{Programming Languages and Frameworks}
We selected Python as the primary programming language for its versatility and extensive libraries. We utilized FastAPI, a Python web framework, for developing high performance APIs. Additionally, we employed Django ORM (Object Relational Mapper) for creating relational database models.
\subsection{Database Systems}
We incorporated multiple database systems to cater to different data requirements. MySQL, a relational database management system, was utilized for structured data storage, while MongoDB, a NoSQL database, was chosen for flexible and unstructured data storage. We also integrated Neo4J, a graph database, to represent complex relationships within the Unani Medicines domain.
\subsection{Libraries and Packages}
Various libraries and packages were installed to support different functionalities. Notable libraries include Scikit-learn for machine learning algorithms, TensorFlow and PyTorch for deep learning models, Pandas for data analysis, and Transformers for natural language processing tasks.
\subsection{Development Tools}
We utilized development tools like customtkinter for building the UTagger desktop application, React for developing the web based CDSS frontend, and version control systems like Git for collaborative development.
\subsection{System Configuration}
We ensured that the environment met the necessary hardware and software requirements for optimal performance. This included setting up compatible operating systems, allocating sufficient computing resources, and installing the required software dependencies.
By carefully configuring the environment, we established a robust and scalable foundation for our project, enabling seamless integration of different components and efficient development of the Online CDSS.
\section{Business Context}
This project is a non-profit, open-source initiative that is focused on advancing the field of Unani Medicines through the development of a comprehensive platform. All APIs and data access will be made publicly available for the benefit of practitioners, patients, and software developers and knowledge engineers. The CDSS component of the platform will provide valuable assistance for healthcare professionals and patients alike, including the storage and management of diagnostic information. In order to sustain and further develop these services, a revenue model will be implemented to enable payment for the use of the CDSS. This will ensure the continued availability and improvement of the system for the benefit of the Unani Medicines community.

\section{Functional Requirements}
Following are the functional requirements of each module:
\subsection{UM Database}
\begin{enumerate}
\item   Access the data in RDB using API. 
\item	Access the data in NoSQL DB using API.
\item	Access the Ontology using API.
\item	Contribute to the RDB by digitization of the structured data.
\item	Contribute to the NoSQL DB by digitization of the unstructured data
\item	Contribute to the Ontology by digitization of the structure data.   
\end{enumerate}

\subsection{AI Inference Engine}
\begin{enumerate}
\item   Access the prebuilt Models for the development of the applications.
\item 	Update the prebuilt Models for improving the quality of models.
\item 	Contribute the new Models for expanding the capabilities of Inference Engine.
\end{enumerate}

\subsection{Online CDSS}
\begin{enumerate}
    \item Sign up for an account.
	\item Log in to my account.
\item	Add symptoms for the diseases.
\item	Identifying the Disease based on the symptoms.
\item	See the details for the relevant disease.
\item   See treatment options for the resultant disease.
\item 	Examining the probability of various diseases to recommend the most suitable treatment.

\end{enumerate}

\section{User Interface Requirements}
\begin{enumerate}
    \item 	Database and Ontology have a Swagger UI Interface for API access.
    \item AI Inference Engine also has a Swagger UI Interface for API access. The models are also hotels on HuggingFace Spaces and Gradio UI is available.
    \item CDSS has a Web UI interface. Users can have access to the capabilities of CDSS using the Web UI interface.

\end{enumerate}

\section{Performance Requirements}
In terms of performance requirements, the UM Database API is designed to operate efficiently even on low-end machines, with no significant performance constraints. Users can easily access the required data, and contribute to the database by following a simple process of verifying their contributions. Similarly, the AI Inference Engine API has no significant performance limitations, but users may require a reliable internet connection to download the AIML models. Running the entire AI inference engine locally may demand high-end computing resources, particularly when developing and training large models on vast amounts of UM data. Online CDSS is a web-based application that does not necessitate any additional resources from users to carry out its operations. Users can access it from any device with a stable internet connection.
\section{Other non-functional attributes}
The application prototype needs to fulfill following non-functional requirements as well:
\begin{itemize}
    \item Security
    \item Compatibility
    \item Reliability
    \item Portability
    \item Extensibility
    \item Reusability
    \item Serviceability
    \item Resource Utilization
\end{itemize}
\clearpage

%% file: Chapters/Chapter4.tex
\chapter{Design of Online CDSS for Unani Medicines } % Write in your own chapter title
\label{Chapter 4}
\lhead{Chapter 4. \emph{Design of Online CDSS for Unani Medicines }}  
This chapter presents the details of the prototype application design. It starts with the proposed architecture that addresses the application’s objectives. The section provides details about three distinct modules of UM Database, Inference Engine, and Online CDSS, each with its own unique architecture.
\section{Architecture}
\label{sec:arc}
\begin{figure}
\includegraphics[width=1\textwidth]{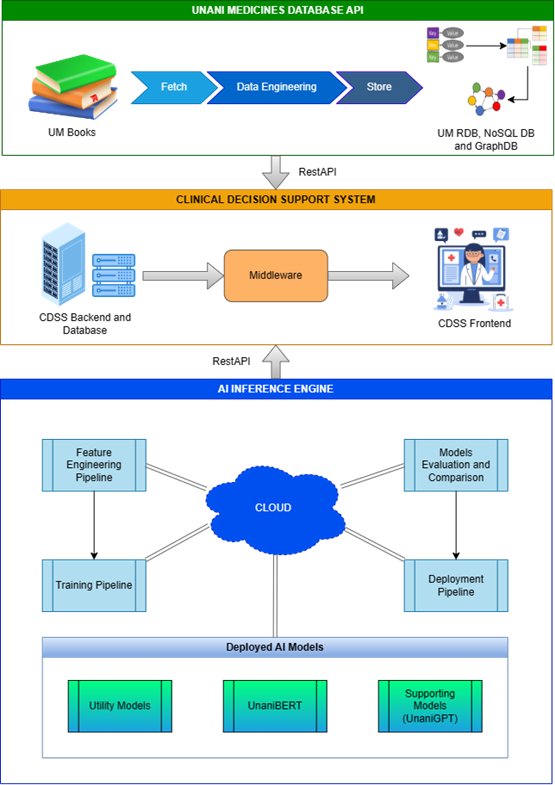}
\rule{35em}{0.5pt}
\caption{High-Level Architecture with modules of  UM Database, CDSS, and AI Engine}
\label{fig:h_arc}
\end{figure}

The High Level Architecture Diagram (Fig \ref{fig:h_arc}) illustrates the 3 core modules of the Unani Medicines (UM) Database, the Online Clinical Decision Support System (CDSS), and the AI Inference Engine. The details of each module are provided in the following.

\subsection{Unani Medicines database for data digitization and storage}
The first module is the UM database, which is responsible for fetching, data engineering, and storing data related to UM principles, diseases, symptoms, and treatment. This module fetches this Unani data from a UM book [6] using a tool called UTagger [11] (which is a desktop application written in Python with a Custom Tkinter library). To tag (or label) the data in the UM books, we have developed a tagging scheme and tagged the data. The tagged file is then used to extract data in a JSON (key, value) pair format. With the keys representing the UM tags and the values representing the corresponding UM data. A RestAPI is provided in UTagger that allows uploading the JSON file to NoSQL, RDB, and GraphDB (Neo4j). The NoSQL database is straightforward to use as it adopts an object model that resembles JSON. Hence, the entire JSON file can be uploaded to NoSQL and queried as required. However, a relational database (RDB) is more complex to handle as it requires the creation of tables and the definition of relations among those tables. JSON files are transformed into RDB via custom Python and Django ORM (Object Relational Mapper). As JSON does not provide such information, we cannot develop an RDB with a single-click API operation. Instead, the API provides functions for manually developing the entire RDB. 
\\ We created our graph database using the RDB. We convert the tables and relations in the RDB into nodes and edges to create the graph database in Neo4j. All of these databases are hosted on a perpetual RDB and NoSQL DB in CleverCloud. And GraphDB is hosted on Neo4J.

\subsection{Technologies Used in the Development of UM Database}
The technologies we used are
\begin{enumerate}
    \item Python (Programming Language) \cite{python}
    \item CustomTkinter (Python Library for Desktop Development, used for UTagger) \cite{tkinter}
    \item FastAPI (Python Library for developing higher performance lightweight asynchronous APIs) \cite{fastapi}
    \item MySQL (RDB) \cite{mysql}
    \item MongoDB (NoSQL DB) \cite{mongodb}
    \item Neo4J (Graph DB Library) \cite{neo4j}
    \item Django ORM (Python Web Development Framework’s Object Relational Mapper for creating RDB Models and performing Database operations) \cite{django}
\end{enumerate}

\subsection{AI Inference Engine}
The AI Inference Engine, integrated with the Online CDSS, utilizes advanced AIML algorithms and trained models to analyze the input data of symptoms entered by the patients (stored in CDSS) and to generate accurate disease diagnoses.
Feature Engineering was carried out on the UM database. This data was further used to engineer NLP training data. We discovered a novel approach to augment our data without sacrificing its quality. We generated NLP prompts (We named it UnaniGPT) based on the previously engineered data (for details see section \ref{section:3.2.2.1}).
Training of the UnaniBERT model (an instance of BioBERT [add reference] for Unani data ) was done using PyTorch. It’s a Multi-Class Classification problem as we have multiple diseases. 
All of these models are deployed on HuggingFace Spaces (cloud) using Gradio.io and the Transformers library. Each model in the AI Inference Engine is a Git Repository. The Gradio SDK provides a containerized version that can be used locally as well.
Users can contribute to these models using pull requests or by joining as contributors directly in the model repository. To unify all the operations in one place for users, RestAPI with Swagger UI is developed, which provides access to all the Unani Medicines models and Spaces from a single location. The API is designed using Python, FastAPI, transformers, and Gradio. It is the user’s responsibility to develop these models.
\subsection{Technologies Used in the Development of AI Inference Engine}
The technologies used are the following:
\begin{enumerate}
    \item Python (Programming Language) \cite{python}
    \item  PyTorch (Python Deep Learning Framework, for NLP. PyTorch, found to be more useful in research and is a much better option with transformers compared to TensorFlow)\cite{pytorch}
    \item Transformers (Python library for HuggingFace transformers) \cite{transformers}
    \item Gradio (Python library that allows you to build and share custom interfaces for your machine learning models using Python code or a simple UI) \cite{gradio}
\end{enumerate}

\subsection{Online Clinical Decision Support System (CDSS)}
The third module is the Online CDSS, which integrates the other two modules of UM database and AI Inference Engine through APIs. CDSS is a prototype web application hosted on a cloud accessible via a browser-based User Interface (UI) (see details in \ref{Chapter 5}). The front end is developed using ReactJS. The backend is developed using Python Django Framework and FastAPI. UM Database is used to develop the Backend Database of CDSS. Patient profiles (EHR) are stored in the CDSS backend database whereas symptoms and disease data are extracted from UM database via API call. 
Online CDSS acts as the interface between patients and UM practitioners. Patients can enter their symptoms into EHR and request online appointments with a UM practitioner (their profile and availability schedules are also stored in the backend database). Whereas UM practitioners can access patient EHR data, and based on symptoms they can receive probable diseases and treatment suggestions from the AI Inference Engine through API calls. That enables UM practitioners to make informed decisions to provide personalized recommendations to patients.
\subsection{Technologies Used in the Development of AI Inference Engine}
The technologies used are the following:
\begin{enumerate}
    \item ReactJS (JavaScript Library for frontend development)
    \item Django (Python Backend Development framework
    \item FastAPI (Lightweight high-performance python library for backend development)
    \item MySQL (Relational Database).
\end{enumerate}

\section{UML and DB Design}
\subsection{UML Use Case Diagram}
\begin{figure}[htbp]
\includegraphics[angle=0]{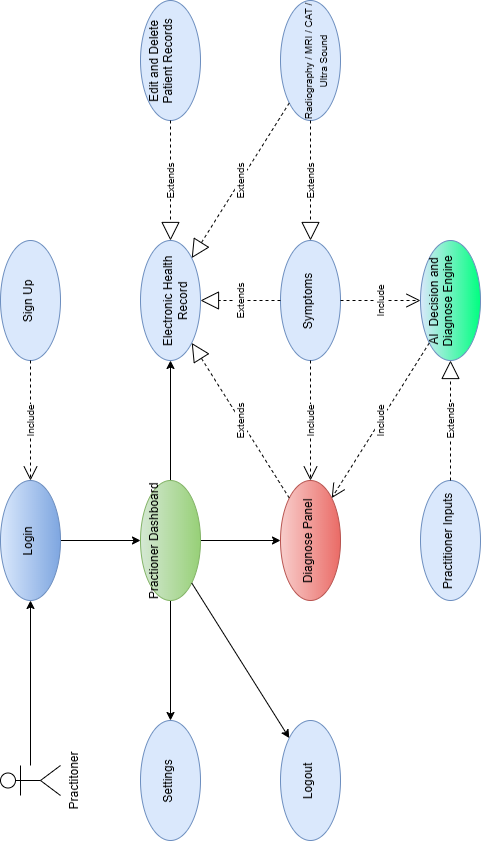}
\rule{35em}{0.5pt}
\caption{Use Cases Diagram - Showing practitioner interactions and use cases.}
\label{fig:uml}
\end{figure}

The use case diagram for the Online CDSS shows the different actors and use cases involved in the application, including the UM practitioner, patient, and the system itself. The use cases include Adding Symptoms, Diagnosing Diseases, and Viewing Prediction Results. The Fig. \ref{fig:uml} illustrates how different  actors interact with the system to perform various tasks and to achieve their goals.
\clearpage

\subsection{UML Sequence Diagram}
\begin{figure}
    \centering
    \includegraphics[height=25cm, width=16cm]{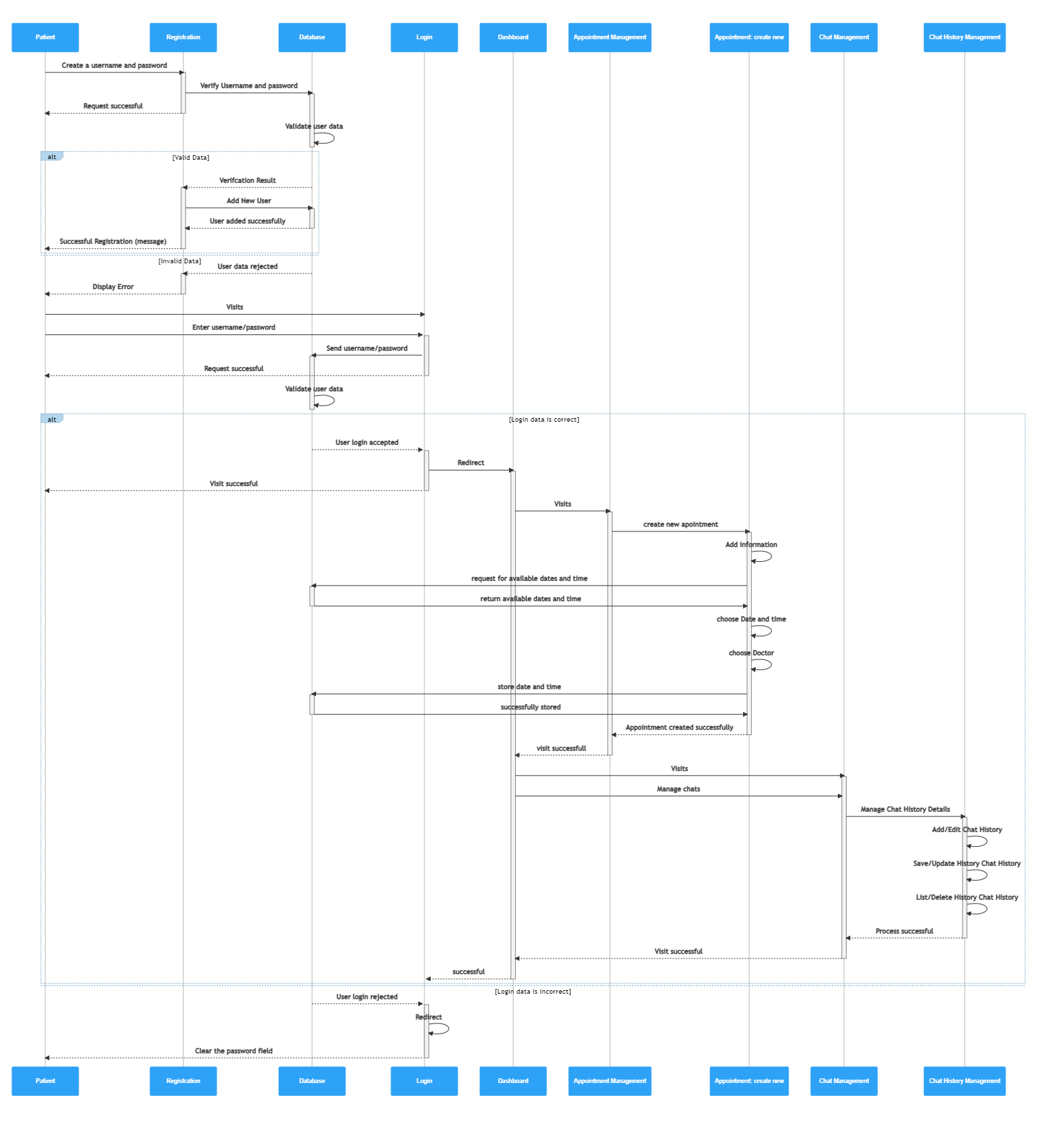}
    \rule{35em}{0.5pt}
    \caption{Sequence Diagram - Showing the dynamic communication flow for patient}
    \label{fig:seqPat}
\end{figure}

Sequence diagrams (See Fig. \ref{fig:seqPat}) for diagnosing a patient’s disease in the online Clinical decision support system. This diagram shows the interactions between the Unani Medicines practitioner, the patient, and the system during the diagnosis process. The sequence of events starts with the practitioner adding the patient’s symptoms to the system and ends with the system providing a diagnosis based on those symptoms. The diagram illustrates the message passing and communication flow between the different components of the system during the diagnosis process.
\begin{figure}[htbp]
   \centering
\includegraphics[height=25cm, width=16cm]{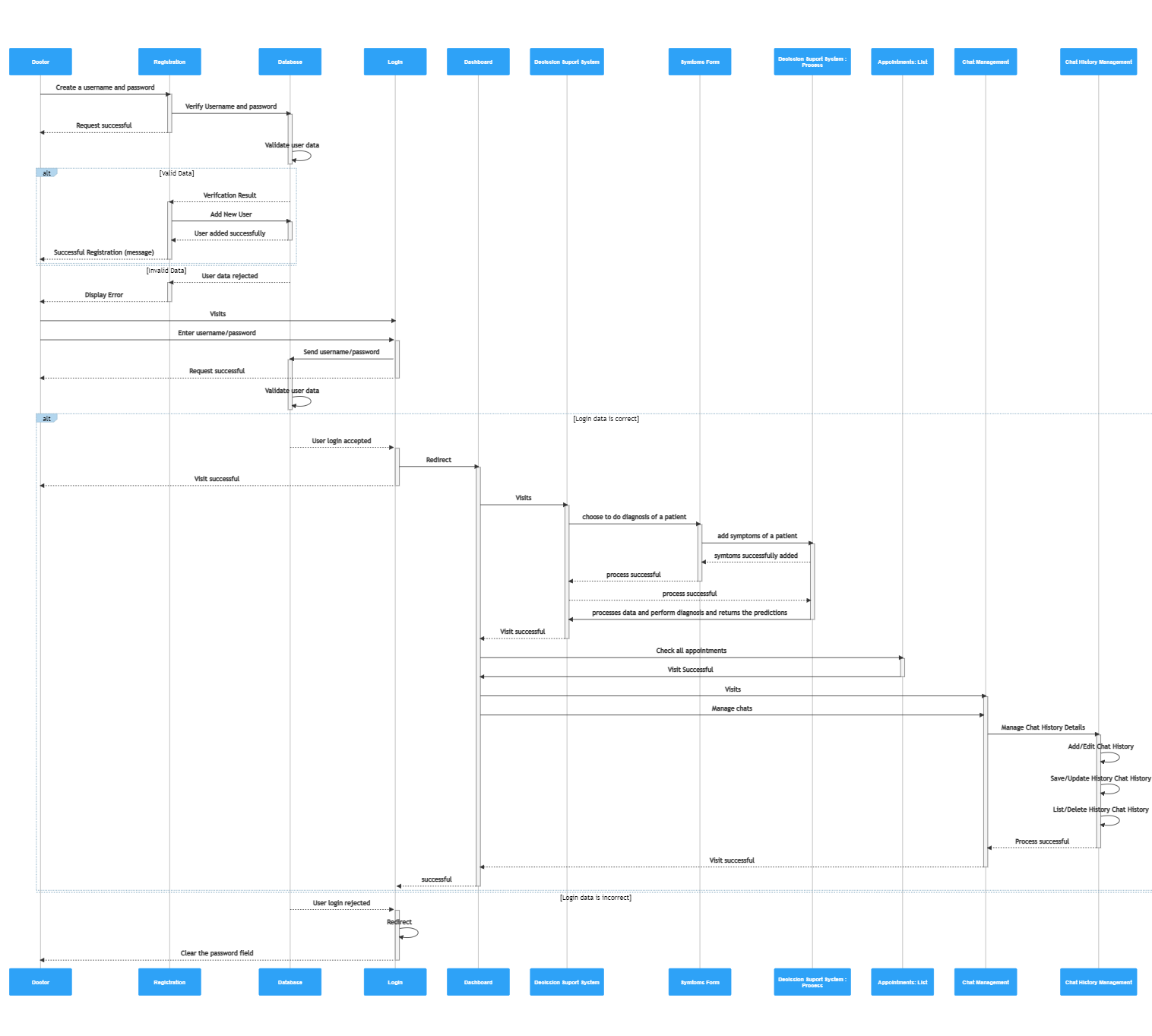}
\rule{35em}{0.5pt}
\caption{Sequence Diagram - Showing the dynamic communication flow for the practitioner}
\label{fig:sqe}
\end{figure}
\clearpage

\clearpage

\subsection{Entity Relation Diagram}
\begin{figure}[htbp]
    \centering
    \includegraphics[height=20cm, width=15cm]{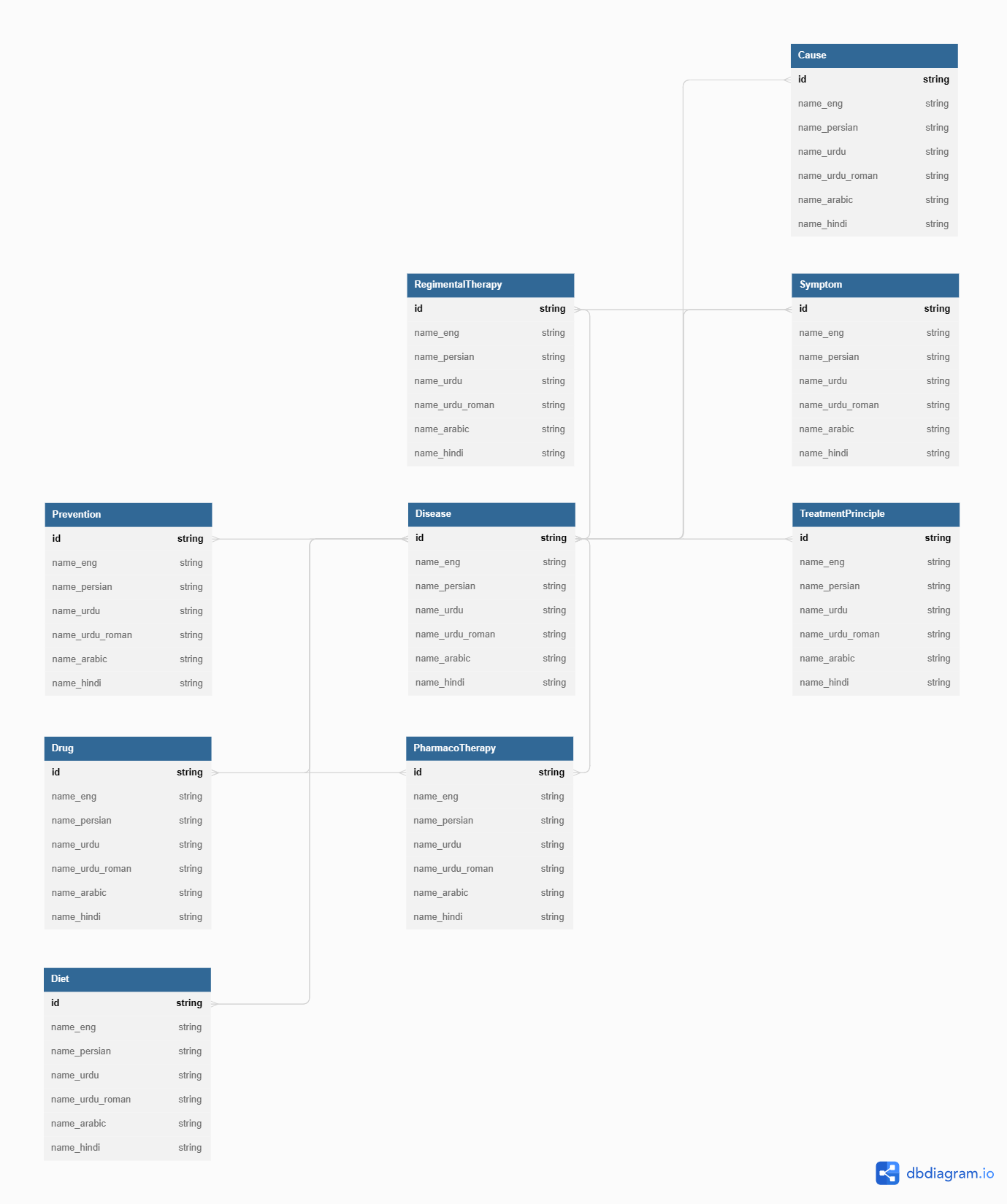}
    \rule{35em}{0.5pt}
    \caption{ERD of Online CDSS showing entities, relationships, and data flow}
    \label{ERD}
\end{figure}

Entity Relationship Diagram (Fig. \ref{ERD}) for the online Clinical decision support system’s database. This diagram shows the different entities involved in the system and the relationships between them. The entities include Patient, Disease, Symptom, Diagnosis, and Practitioner. The diagram illustrates how the entities are connected through their relationships, such as the Patient having multiple Symptom instances and receiving multiple Diagnosis instances. The ERD provides a visual representation of the database schema and helps in understanding the data flow and relationships between different entities.

\begin{figure}
    \centering
    \includegraphics[height=20cm, width=15cm]{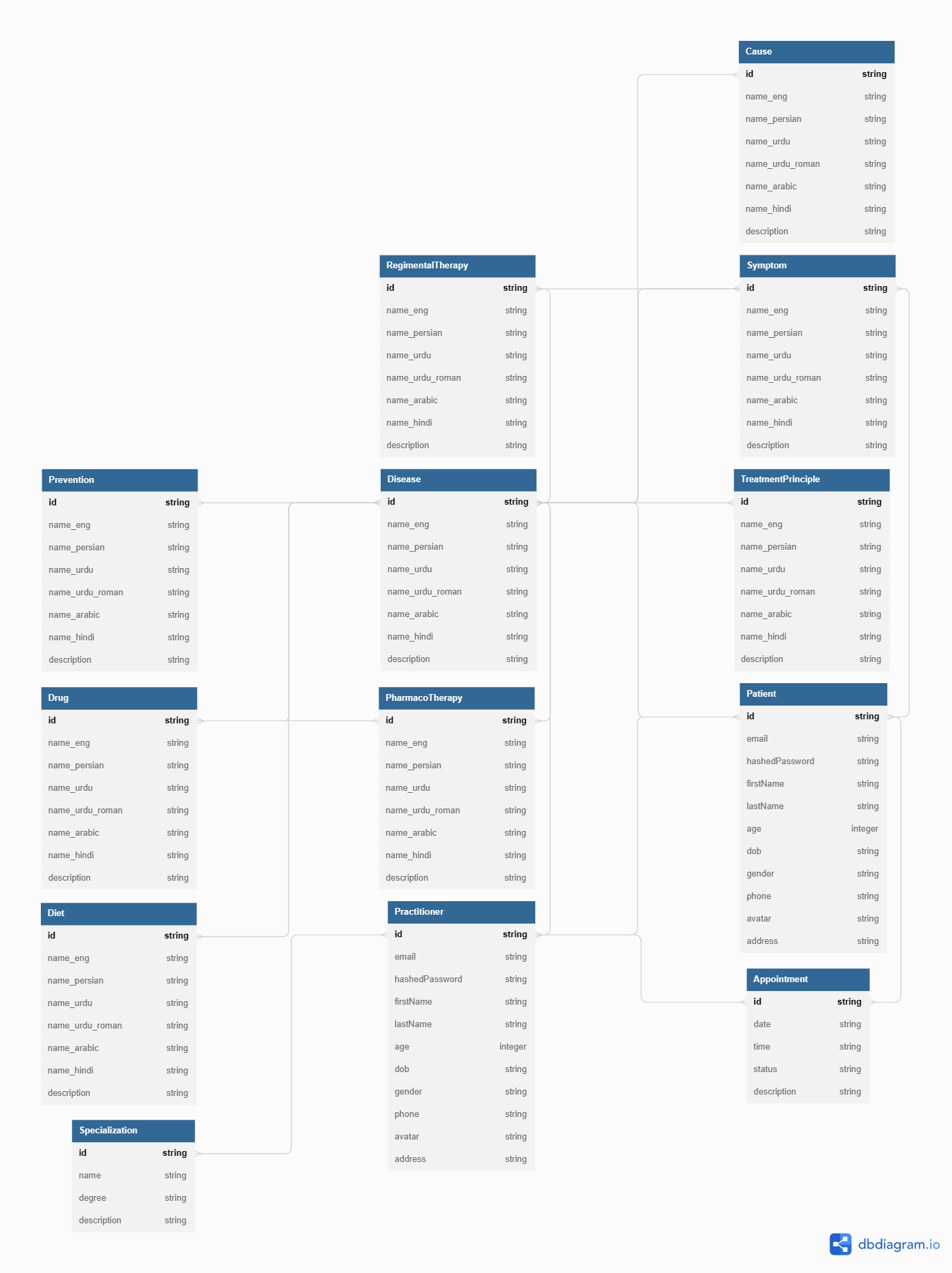}
    \rule{35em}{0.5pt}
    \caption{ERD showing the entities, relationships, and data flow within the system}
    \label{CDSS_ERD}
\end{figure}
\clearpage

\section{SWRL Rules Usage}
Semantic Web Rule Language (SWRL) rules extend the capabilities of decision trees by formalizing complex relationships and logical deductions. These rules are used to describe and encode the conditions and outcomes that guide the decision-making process. By defining rules using SWRL, we enhanced the decision tree's expressiveness, enabling it to handle intricate decision logic that may involve multiple variables and relationships. This integration bridges the gap between the intuitive "if-else" structure of decision trees and the semantic representation offered by SWRL, ultimately allowing for more sophisticated and context-aware decision-making processes.

\section{Example Clinical Rules in SWRL}
We referred to the Standard Unani Treatment Guidelines for Common Diseases (Volume-1) \cite{ccrum2014standard} as  a source for disease and treatment related clinical rules and coded these into Semantic Web Rule Language (SWRL). These rules are essentially a formal representation of the descriptive information contained within the book. Few examples of the diseases from the book are given in the following section that we used for developing clinical rules in SWRL.

\subsection{Migraine (Shaqīqa)}
\subsubsection{Introduction - Symptoms}
\begin{enumerate}
    \item It is the type of headache in which only one half of head is afflicted with pain.1 Sometimes it involves the whole head.
    \item  It is characterized by episodic throbbing pain in only one half of head (especially in Damawī type) due to lesser quantity of causative matter. There may be relief in pain when the throbbing artery is pressed.1 The headache is frequently accompanied with Tanīn (Tinnitus) and Ghasayān (Nausea).
\end{enumerate}
\begin{equation}
    \boxed{
  \begin{aligned}
      & \textbf{Symptoms}(?p,\text{half\_head\_episodic\_throbbing\_pain}), \nonumber \\ & \textbf{Symptoms}(?p,\text{whole\_head\_sometimes}) \rightarrow \textbf{hasDisease}(?p, \text{Migraine}) \\
  \end{aligned}
  }
\end{equation}

\subsubsection{Introduction - Causes}
It is caused by Bukharaat (Vapours), arising towards the head from the 	body or by Akhlaat Harra (Hot humours) or Baarida (Cold humours). In 	this case a relatively less quantity of causative matter intervenes.
\begin{equation}
\boxed{
\begin{aligned}
& \textbf{Causes}(?p, \text{Vapours\_arising\_towards\_head\_from\_body}), \\
& \textbf{Causes}(?p, \text{hot\_humours}), \nonumber \\
& \textbf{Causes}(?p, \text{cold\_humours}) \rightarrow \textbf{hasDisease}(?p, \text{Migraine})
\end{aligned}
}
\end{equation}

\subsubsection{Principles of treatment (Usūl-i ‘Ilāj )}
\begin{enumerate}
    \item Taskīn-i Dard (Analgesia)
    \item Istifrāgh (Evacuation) of causative Khilt (Humour)
    \item Taqwiyat-i Dimāgh (Toning up of brain)
\end{enumerate}
\begin{equation}
\boxed{
\begin{aligned}
& \textbf{Treatment}(?p, \text{Analgesia}), \nonumber \\
& \textbf{TreatmentPrinciples}(?p, \text{Causative\_humouur\_Evacuation}), \\
& \textbf{TreatmentPrinciples}(?p, \text{ToningUp\_Of\_Brain}) \rightarrow \textbf{hasDisease}(?p, \text{Migraine})
\end{aligned}
}
\end{equation}

\subsubsection{Regimenal therapy (Ilāj bi’l-Tadbīr)}
\begin{enumerate}
    \item Fasd (Bloodletting) in case of Damawī type.
    \item Ishāl (Purgation) in case of Safrāwī, Sawdāwī and Balghamī types.
    \item Natūl (Irrigation) Hār/Bārid
\end{enumerate}
\begin{equation}
\boxed{
\begin{aligned}
& \textbf{RegimentalTherapy}(?p, \text{Bloodletting}), \nonumber \\
& \textbf{RegimentalTherapy}(?p, \text{Purgation}), \\
& \textbf{RegimentalTherapy}(?p, \text{Irrigation}) \rightarrow \textbf{hasDisease}(?p, \text{Migraine})
\end{aligned}
}
\end{equation}
\subsubsection{Prevention/Precaution (Tahaffuz)}
Grief \& Sorrow to be avoided.
\begin{equation}
\boxed{
\begin{aligned}
& \textbf{Prevention}(?p, \text{Grief\_Avoidance}), \nonumber \\
& \textbf{Prevention}(?p, \text{Sorrow\_Avoidance}) \rightarrow \textbf{hasDisease}(?p, \text{Migraine})
\end{aligned}
}
\end{equation}
\subsection{Insomnia(Sahar)}
\subsubsection{Introduction - Symptoms}
\begin{enumerate}
    \item It is the excess of wakefulness and inability to fall asleep as long as 
	desired for a normal person.
    \item It is characterized by disorientation, feeling of weightlessness in 
        the head, dryness of eyes, tongue and nostrils (may be moist when caused by Rutūbat Būraqiyya), excessive thirst and burning sensation in the eyes.
\end{enumerate}
\begin{equation}
\boxed{
\begin{aligned}
& \textbf{Symptoms}(?p, \text{excess\_wakefulness}), \nonumber \\
& \textbf{Symptoms}(?p, \text{inability\_to\_fall\_asleep\_for\_desired\_time}), \\
& \textbf{Symptoms}(?p, \text{disorientation}), \\
& \textbf{Symptoms}(?p, \text{feeling\_of\_head\_weightlessness}), \\
& \textbf{Symptoms}(?p, \text{eyes\_tongue\_nostrils\_dryness}), \\
& \textbf{Symptoms}(?p, \text{excessive\_thirst}), \\
& \textbf{Symptoms}(?p, \text{eyes\_burning\_sensation}) \rightarrow \textbf{Disease}(?p, \text{Insomnia})
\end{aligned}
}
\end{equation}
\subsubsection{Introduction - Causes}	
It is caused by the predominance of Harārat Sāda (Simple heat), Yubūsat (Dryness), Safrā’ (yellow bile), Sawdā’ (Black bile), Balgham Shor or deep seated Rutūbat Būraqiyya (Alkaline secretion) in the brain, pain and stress
\begin{equation}
\boxed{
\begin{aligned}
& \textbf{Causes}(?p, \text{simple\_heat\_predominance}),  \nonumber \\
& \textbf{Causes}(?p, \text{dryness}), \\
& \textbf{Causes}(?p, \text{yellow\_bile}), \\
& \textbf{Causes}(?p, \text{black\_bile}), \\
& \textbf{Causes}(?p, \text{deep\_seated\_AlkalineSecretion\_in\_brain}), \\
& \textbf{Causes}(?p, \text{pain}), \\
& \textbf{Causes}(?p, \text{stress}) \rightarrow \textbf{Disease}(?p, \text{Insomnia})
\end{aligned}
}
\end{equation}

\subsubsection{Principles of treatment (Usul-ilaj )}
\begin{enumerate}
    \item Tartīb (Producing moistness)
\item Tadhīn (Producing moistness through oils)
\item Taskīn-i Dard (Analgesia) in case of pain
\item Sukūn-i Jismānī o Nafsānī (Physical \& mental rest)
\end{enumerate}
\begin{equation}
\boxed{
\begin{aligned}
& \textbf{Disease}(?p, \text{Insomnia}) \rightarrow \nonumber \\
& \textbf{TreatmentPrinciples}(?p, \text{moist\_Production}), \\
& \textbf{TreatmentPrinciples}(?p, \text{Analgesia}), \\
& \textbf{TreatmentPrinciples}(?p, \text{Physical\_}\&\text{mentalRest}), \\
& \textbf{TreatmentPrinciples}(?p, \text{extremitiesMassage}), \\
& \textbf{TreatmentPrinciples}(?p, \text{Irrigation})
\end{aligned}
}
\end{equation}

\subsubsection{Regimenal therapy (Ilaj bil-Tadbir)}
\begin{enumerate}
    \item Natūl (Irrigation)
    \item Hammām Mu‘tadil
    \item Dalk-i Atrāf (Massage on the extremities)
\end{enumerate}
\begin{equation}
\boxed{
\begin{aligned}
& \textbf{Disease}(?p, \text{Insomnia}) \rightarrow \nonumber \\
& \textbf{RegimentalTherapy}(?p, \text{Irrigation}), \\
& \textbf{RegimentalTherapy}(?p, \text{Massage\_on\_Extremities})
\end{aligned}
}
\end{equation}

\subsubsection{Prevention/Precaution (Tahaffuz)}
Indigestion, Fikr (Mental stress), Kasrat-i Jimā‘ (Excessive coitus), Ta‘b 	(Exertion), Gham o Alam (Grief \& Sorrow), Afkār Mushawwisha (Apprehensions) and factors causing Yubūsat (Dryness) are to be avoided.
\begin{equation}
\boxed{
\begin{aligned}
& \textbf{Disease}(?p, \text{Insomnia}) \rightarrow \nonumber \\
& \textbf{Prevention}(?p, \text{Grief\_Avoidance}), \\
& \textbf{Prevention}(?p, \text{Excessive\_Coitus}), \\
& \textbf{Prevention}(?p, \text{Exertion}), \\
& \textbf{Prevention}(?p, \text{Dryness}), \\
& \textbf{Prevention}(?p, \text{Apprehensions})
\end{aligned}
}
\end{equation}

%% file: Chapters/Chapter5.tex
\chapter{Implementation of Online CDSS for Unani Medicines} % Write in your own chapter title
\label{Chapter 5}
\lhead{Chapter 5. \emph{Implementation of Online CDSS for Unani Medicines}}
Overall, this chapter provides an insight into the implementation details of the CDSS, showcasing its user interface, application server, backend components, and modules that contribute to its functionality in assisting Unani Medicines practitioners in diagnosing diseases and providing treatment recommendations to patients.

\section{Overview}
A web-based prototype application will be used as Online CDSS for Unani Medicines practitioners to diagnose diseases based on the symptoms given by the patients. The system is made to help UM practitioners in making diagnoses and advising patients’ best courses of treatment. The practitioners will also be able to enter symptoms, look at severity indices for various diseases, and issue alerts for disease prevention.
\\ React, Next.js, Python, and other cutting-edge web technologies were used for application implementation. For simplicity of access and scalability, the application will be hosted online. The CDSS UI, architecture, and modules are covered in the following section.
\section{Modules}The system is made up of a number of modules \ref{fig:x PS} that interact with one another to give consumers the functionality they require. The system's primary modules are as follows:
\begin{enumerate}
    \item User Management: The user management module is in charge of overseeing system users. It enables users to 
        \begin{enumerate}
            \item Register for an account.
            \item Login to the account. 
            \item Manage their account settings.
        \end{enumerate}  
    \item Disease diagnostic: Based on the symptoms given by the patients, the disease diagnostic module is in charge of making a diagnosis of the disorders. In order to analyse the symptoms and offer a diagnosis report, it employs machine learning techniques.
    \item Treatment Management: The treatment management module is in charge of overseeing the available treatments for the disorders that have been identified. It offers the required therapy options to the practitioners.
\end{enumerate}

\section{CDSS Components}
\subsection{User interface} The user interface is the system's front-end element and it offers the panels and forms people need to engage with it. React was used to build the user interface, which has a responsive and approachable design.
\subsubsection{CDSS Dashbaord}
\begin{figure}[htbp]
\centering
\includegraphics[height=10cm, width=15cm]{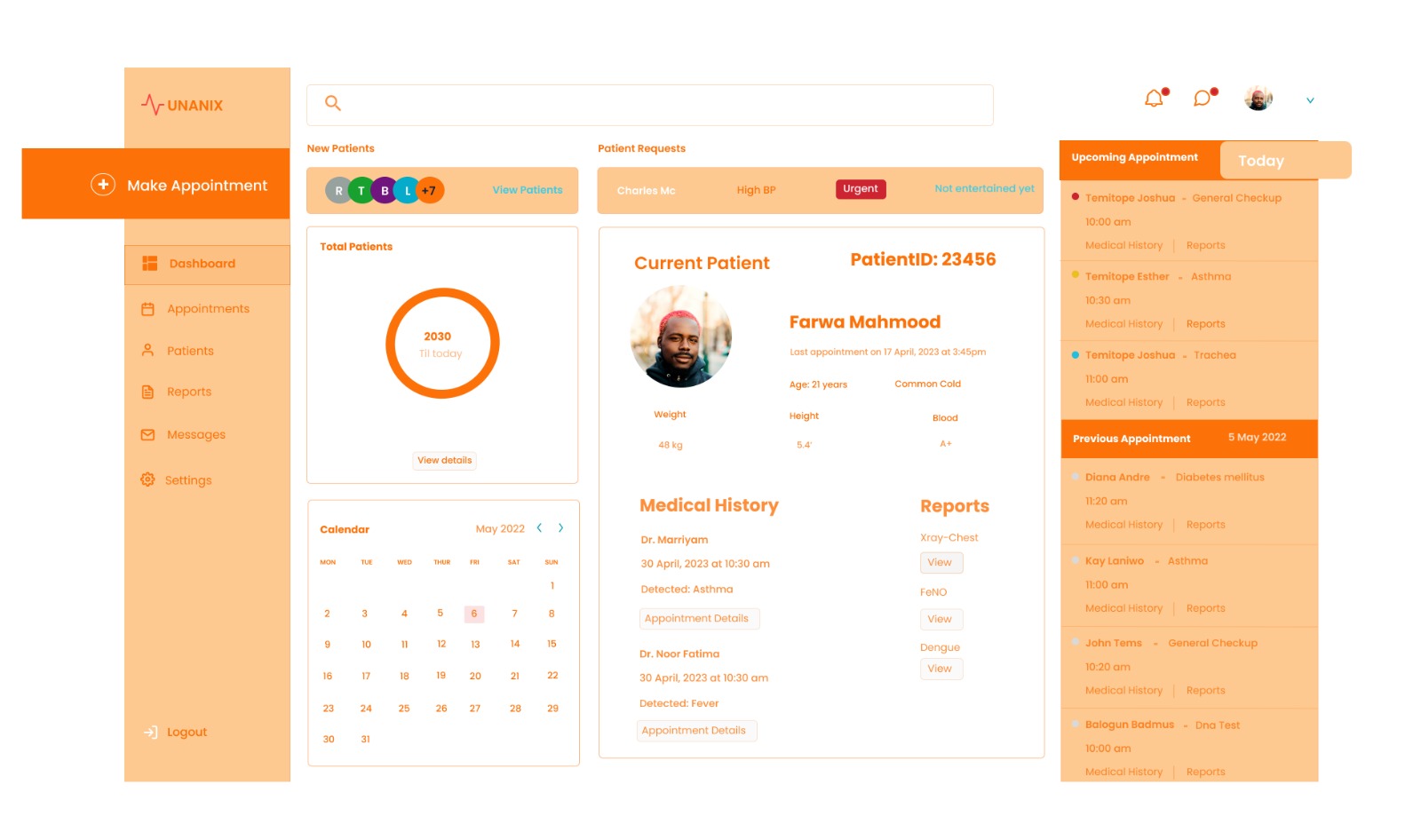}
\caption{Screenshot of Dashboard of Online CDSS for Unani Medicines}
\label{fig:x DashBoard}
\end{figure}

Ayesha, a 22-year-old woman, visits a Unani medicine practitioner with symptoms of a running nose and a headache. The practitioner uses the Online Clinical Decision Support System (CDSS) to diagnose her condition and suggest appropriate treatment.
Upon logging in, the practitioner is greeted by the main dashboard (Fig. \ref{fig:x DashBoard}) displaying an overview of the system's user interface. The dashboard presents charts and graphs depicting statistical distributions of patients' appointments, including the distribution of male and female patients. It also lists patients assigned to the practitioner. This user-friendly interface allows the practitioner to quickly grasp the system's prediction results and patient data.

\clearpage

\subsubsection{Patient Symptoms}
\begin{figure}[htbp]
\centering
\includegraphics[height=7cm, width=15cm]{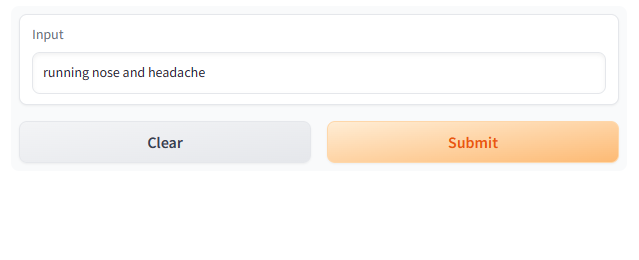}
\caption{Screenshot for Entering Patient symptoms}
\label{fig:x PS}
\end{figure}

The patient symptoms screenshot (Fig. \ref{fig:x PS}) shows the page where the Unani Medicines practitioner can enter the patient’s symptoms into the system.
The practitioner navigates to the "Patient Symptoms" section (Fig. \ref{fig:x PS}) to enter Ayesha's symptoms. The UI includes a textbox where the patient’s symptoms can be added in the form of a defined sentence. In a textbox, the practitioner types in the symptoms: "running nose" and "headache." The interface's simplicity ensures efficient and accurate input of patient symptoms. This information is crucial for the AI Inference Engine to process and diagnose Ayesha's disease.

\clearpage

\subsubsection{AI Inference Engine Diagnosis}
\begin{figure}[htbp]
\centering
\includegraphics[height=10cm, width=15cm]{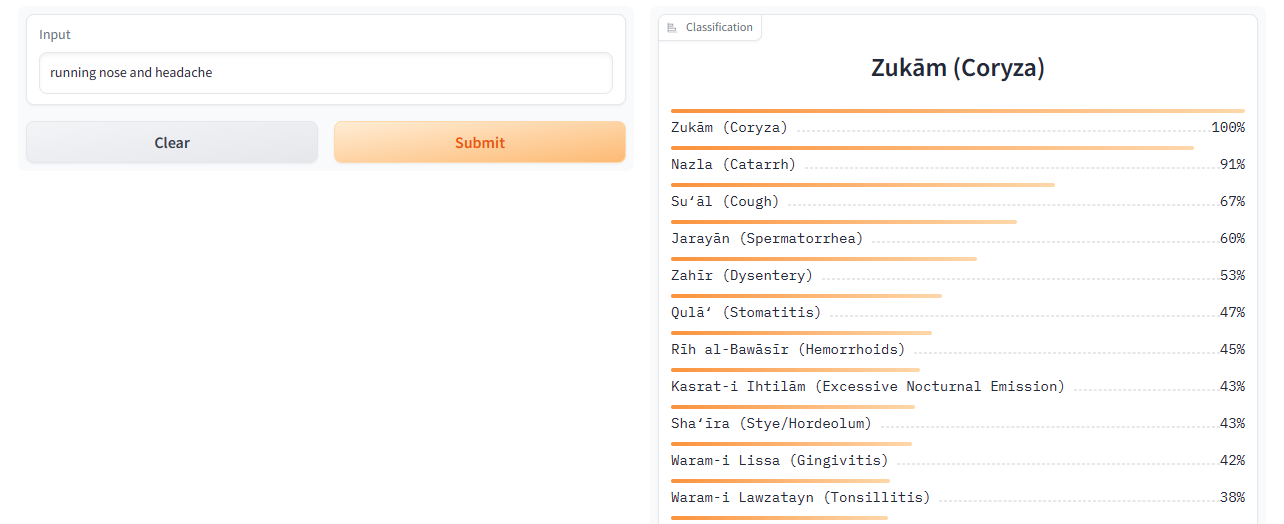}
\caption{Screenshot of Predicted Diseases On Right-Hand side by AI Inference Engine}
\label{fig:x AIInfEng}
\end{figure}

After inputting Ayesha’s symptoms, the practitioner proceeds to the "Predicted Diseases" page (Fig. \ref{fig:x AIInfEng}) where the AI Inference Engine has generated a list of potential diseases based on the symptoms. Each entry includes the system's confidence level and additional details. The practitioner reviews the list and identifies the most likely diagnosis. In Ayesha's case, the diagnosis is "zukam" (Flu).
\clearpage
\subsubsection{Treatment}
To determine the appropriate treatment, the practitioner accesses the "Treatments" section (Fig. 5.4). The UI displays a list of recommended treatments for Ayesha’s diagnosed disease, "zukam." The practitioner sees the following suggestions:
\begin{itemize}
    \item Hot fomentation
    \item Steam inhalation
\end{itemize}
Regimental Therapy:
\begin{itemize}
    \item "Takmīd Hār" (Hot fomentation) for "Zukām Bārid" (Cold Flu)
    \item "Fasd" (Bloodletting) for "Zukām Hār" (Hot Flu)
    \item "Hammām" (Bath) for both "Zukaām Hār" (Hot Flu) and "Bārid" (Cold Flu)
\end{itemize}
These recommendations are informed by the AI Inference Engine's analysis of the symptoms and disease pattern in Unani medicine. The practitioner can confidently select the most suitable treatments for Ayesha's condition, enhancing the quality of care.
\\ In this scenario, the CDSS effectively aids the Unani medicine practitioner in diagnosing Ayesha's condition and proposing tailored treatments. The integration of advanced technology with traditional Unani medicine principles improves accuracy, efficiency, and patient outcomes.

\begin{figure}[htbp]
\centering
\includegraphics[height=10cm, width=15cm]{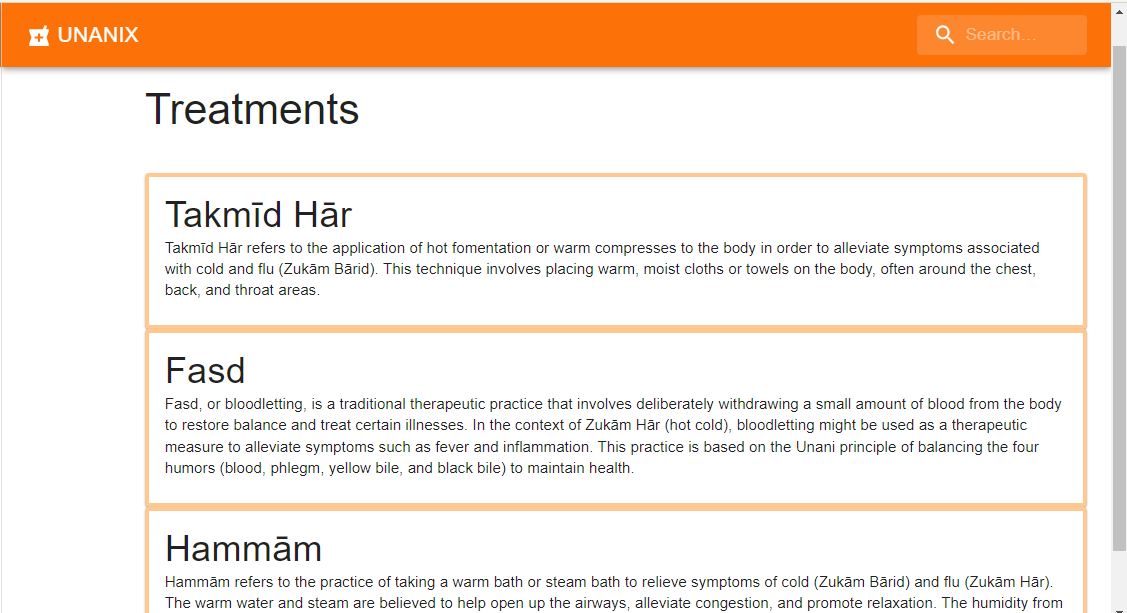}
\caption{Screenshot of Dashboard for Suggested Treatment}
\end{figure}

\clearpage
\subsection{Application Server} 
The application server is the backbone of the CDSS, responsible for handling business logic and API queries. It interacts with the AI Inference Engine API to provide diagnosed diseases based on the symptoms (provided by patient or practitioner). The application server acts as an intermediary between the user interface and the AI Inference Engine, processing the data input by the user and sending it to the AI Inference Engine for analysis. It then retrieves the results and 
presents them to the user in a format that is easily understandable. The application server is a crucial component of the CDSS, as it facilitates the flow of data and allows for seamless communication between the user and the AI Inference Engine.

\subsubsection{Database Server} The database server is responsible for managing the storage and retrieval of patient data and diagnosis reports. It serves as a central repository for all the data needed by the CDSS. The Database and Ontology API is utilized to provide the UM data to the CDSS, enabling it to access and retrieve the relevant data it requires for analysis. The database server stores patient’s medical data in a secure and organized manner, ensuring that it is readily accessible to the application server when required. The use of a database server is a critical component of the system, as it ensures the efficient management and retrieval of patient data, making it easier for healthcare professionals to provide accurate diagnoses and treatment recommendations.

\section{Summary}
This chapter provides an overview of the implementation of the online Clinical Decision Support System (CDSS) for Unani Medicines practitioners. The chapter discusses the components of the CDSS, starting with the user interface which provides a user-friendly and intuitive design for practitioners to interact with the system. The UI includes a dashboard that displays statistical distributions of patients with appointments, patient symptoms, predicted diseases based on symptoms, and suggested treatments for diagnosed diseases.
\\The application server serves as the backbone of the system, handling business logic and API queries. It interacts with the AI Inference Engine API to provide medical diagnoses and retrieves results for presentation to the user. The database server manages the storage and retrieval of patient data and diagnosis reports, ensuring efficient management and accessibility of data.

%% file: Chapters/Chapter6.tex
\chapter{Conclusion and Future Work} % Write in your own chapter title
\label{Chapter 6}
\lhead{Chapter 6. \emph{Conclusion and Future Work}}  
This chapter concludes the study and provides an overview of potential areas for future research that incoming students may wish to explore to extend the scope and impact of the project.
\section{Summary of the Project}
The Online CDSS is a web-based application designed to assist Unani Medicines practitioners in diagnosing patients’ diseases accurately and efficiently (see fig \ref{fig:h_arc} for details). The CDSS provides a user-friendly interface for practitioners to enter patient symptoms, and then utilizes advanced data analytics techniques to generate a list of likely diagnoses and suggest treatments for the diagnosed disease. The project team used a strong technology stack that includes React, FastAPI, and MySQL to develop the CDSS.The project consists of three modules: UM database, an AIInference Engine and a CDSS. In the first module of UM database, Unani data in books were tagged using a custom Python desktop app called UTagger and stored in various databases using FastAPI, MySQL, MongoDB, and Neo4J. The second module of Inference Engine involves classifying diseases based on symptoms and causes using various techniques including Decision Trees, Deep Learning, and Natural Language Processing (NLP).
\section{Conclusion}
In conclusion, the prototype application created for this project offers a strong starting point for further work. To improve the functionality and usability of the application, numerous modules can be extended. Some potential future research directions include multilingual assistance, telemedicine integration, wearable device integration, data analytics and machine learning, and EHR integration.
\section{Future Work}
The prototype application is just a proof of concept, during its implementation a number of ideas emerged that have been listed in the following, so upcoming teams can implement them.	
\subsection{Multi-Lingual Support }  
Currently, the prototype application is only fully capable of supporting the Unani Medicines data in English language. However, Unani Medicines is still widely used in various countries and regions where its printed and handwritten books and materials exist in various languages like Urdu, Persian, Arabic, Hindi, Sanskrit etc. To broaden the system’s accessibility and usability, future work could focus on extending language support. This can be accomplished by adding a language detection mechanism, allowing users to select their preferred language. Some work done by us in this regard is by providing Google’s T5 model for translation of data and making the databases multilingual but it’s in the initial stages. Additionally, the system’s content can be translated into multiple languages by a team of professional translators.
\subsection{Generative Algorithms} 
The current system was primarily developed as a CDSS, and the BioBERT model was fine-tuned (and named as UnaniBERT) on an intermediate-sized dataset to classify diseases based on their symptoms and causes. Although we provided a basic solution for integrating a generative model by fine-tuning BioGPT (we named it UnaniGPT), it has limited capability. Future software developers and knowledge engineers have the opportunity to refine these generative algorithms and develop their own versions, such as UnaniGPT and UnaniBERT, which could potentially enhance the system’s overall performance.
\subsection{Conversational AI}
The current version of UnaniBERT, which is a fine-tuned variant of BioBERT, is currently limited to solving classification problems. However, with the development of large language models (LLMs), we can expect to see more sophisticated solutions emerge that utilize multiple models to create conversational AI with human-like capabilities. These advanced models can potentially enhance the system’s ability to engage in dialogue with users, providing more personalized and intuitive recommendations and solutions for their specific needs.
\subsection{Question Answering Models}
We have developed initial level querying models that can effectively respond to user queries from both textual documents and structured tables. These models offer a direct approach to querying data without the need for pre-processing or structuring the data. Furthermore, such models can be refined and customized to cater to the unique characteristics of Unani Medicines data. These models provide software developers and knowledge engineers with a straightforward way of accessing both structured and unstructured data, facilitating the development of innovative solutions in the field.
\subsection{Integration with Telemedicine}
After COVID19 pandemic, there has been an increasing adoption of telemedicine as a means of providing remote healthcare services. In light of this, there is potential for the Unani Medicines solution to be integrated with popular telemedicine platforms such as Zoom or Skype, Microsoft Teams, Google Meet in the future. This would enable doctors to remotely consult with their patients using the UM system as a diagnostic tool. Additionally, a chatbot can be developed and integrated into the system to allow patients to input their symptoms and receive preliminary diagnoses before consulting with a medical professional. This would provide patients with greater access to healthcare services and potentially reduce the burden on healthcare systems.
\subsection{Integration with Wearable Devices}
In order to further improve the capabilities of the system, integration with wearable devices such as fitness trackers and smartwatches can be considered. By allowing patients to input information related to their physical activity, heart rate, and other vital indicators, doctors can receive a more comprehensive understanding of the patient’s health status, enabling more accurate diagnoses. Additionally, the system can be equipped with continuous monitoring functionality to constantly track patients’ health status, alerting medical professionals of any concerning changes. This would provide a more proactive approach to healthcare management, potentially leading to improved patient outcomes.
\subsection{Integration with EHR}
Integration of Electronic Health Records (EHRs) and Electronic Medical Records (EMRs) into the system is a promising avenue for future research and enhancement. By incorporating these records, practitioners would have access to patients' complete medical histories, including test results, past diagnoses, treatments, medications, and laboratory data. This integration would significantly aid in making more informed decisions and improve the diagnostic capabilities of the Unani Medicines solution. Furthermore, leveraging machine learning algorithms on the integrated records can identify patterns and trends, leading to more accurate and efficient recommendations for diagnosis and treatment.
\subsection{Upcoming Software Solutions}
In the ever-evolving landscape of technology, software solutions play a pivotal role in transforming industries and addressing complex challenges. As organizations strive to stay ahead of the curve and meet the changing needs of their customers, it becomes crucial to explore upcoming software solutions that offer innovative approaches and cutting-edge capabilities.
Upcoming software solutions represent the next wave of advancements and possibilities in various domains, ranging from healthcare and finance to communication and entertainment. These solutions leverage emerging technologies, such as artificial intelligence, machine learning, blockchain, and Internet of Things, to revolutionize processes, enhance user experiences, and drive efficiency.
The purpose of this chapter is to shed light on some of the upcoming software solutions that hold significant potential for shaping the future of the UM industry and CDSS. These solutions have the power to address existing challenges, uncover new opportunities, and pave the way for disruptive innovations.
\subsection{Data Privacy and Security}
As the Unani Medicines solution deals with sensitive patient information, ensuring robust data privacy and security measures is crucial. Future work should focus on implementing advanced security protocols to protect patient data from unauthorized access and potential cyber threats. This may include encryption techniques, access controls, audit logs, and regular security audits to identify and address vulnerabilities. Compliance with relevant data protection regulations, such as the General Data Protection Regulation (GDPR), should also be considered to maintain the trust and confidentiality of patient information.
\subsection{Continuous Improvement and User Feedback}
To enhance the effectiveness and usability of the Unani Medicines solution, it is important to establish a feedback loop with users, including practitioners and patients. Collecting feedback on the system's performance, user experience, and suggestions for improvement can provide valuable insights for further development. Regular user surveys, interviews, and usability testing sessions can be conducted to gather feedback and identify areas for refinement. This user-centric approach will help ensure that the software solution continues to meet the evolving needs of its users and remains relevant in the healthcare industry.
\subsection{Integration with External Knowledge Sources}
To enrich the knowledge base of the CDSS and improve its diagnostic capabilities, integration with external knowledge sources can be explored. This may include leveraging medical literature databases, research articles, clinical guidelines, and expert opinions. By integrating these external knowledge sources, the system can stay up-to-date with the latest medical advancements, incorporate evidence-based practices, and provide more accurate and reliable diagnoses and treatment recommendations.
\subsection{Mobile Application Development}
Developing a mobile application version of the Unani Medicines solution can significantly enhance its accessibility and convenience for both practitioners and patients. A mobile app would allow users to access the system's features and functionalities on their smartphones or tablets, providing on-the-go access to diagnostic support and treatment recommendations. The mobile app can also leverage device capabilities such as push notifications and location services to provide personalized reminders and alerts for medication adherence and follow-up appointments.
\subsection{Longitudinal Patient Monitoring}
Expanding the CDSS's capabilities to include longitudinal patient monitoring can provide valuable insights into the effectiveness of treatments and the progression of diseases over time. By integrating with long-term monitoring devices and collecting regular patient data, the system can track changes in symptoms, vital signs, and other relevant parameters. This longitudinal data can be used to refine the system's algorithms, improve accuracy in predicting disease progression, and tailor treatment recommendations based on individual patient profiles.